%% file: main.tex
\ifdefined\pdfminorversion\pdfminorversion=7\fi
\documentclass{article}
\usepackage{iclr2027_conference,times}
\iclrfinalcopy
\usepackage[utf8]{inputenc}
\usepackage[T1]{fontenc}
\usepackage{hyperref}
\hypersetup{hidelinks,pdftitle={Dual-Scale Condition-Shift Editing for Robust Real-World Image Demoireing},pdfauthor={Shiyang Zhou; Haijin Zeng; Congpei Qiu; Zhengcen Li; Ping Chen; Xiangming Wang; Zheng Li; Yongyong Chen; Jingyong Su}}
\usepackage{url}
\usepackage{booktabs}
\usepackage{amsfonts}
\usepackage{amsmath}
\usepackage{amssymb}
\usepackage{nicefrac}
\usepackage{microtype}
\usepackage{xcolor}
\usepackage{graphicx}
\usepackage{multirow}
\usepackage{enumitem}
\usepackage{subcaption}
\usepackage{placeins}
\usepackage{wrapfig}
\usepackage{array}
\usepackage{tabularx}
\usepackage{flafter}

\newcommand{\papertablestyle}{\normalsize\setlength{\tabcolsep}{4pt}\renewcommand{\arraystretch}{1.05}}
\newcolumntype{Y}{>{\raggedright\arraybackslash}X}
\title{Dual-Scale Condition-Shift Editing for \\Robust Real-World Image Demoir\'eing}

\input{authors}

\begin{document}
\raggedbottom

\maketitle
\fancyhead{}
\vspace{-12pt}
\begingroup
\centering
\setlength{\parskip}{2pt}
{\normalsize\hypersetup{colorlinks=true,urlcolor=blue}\urlstyle{same}
\textbf{Code:} \href{https://github.com/Clausy9/DemoireEdit}{\nolinkurl{github.com/Clausy9/DemoireEdit}}\qquad
\textbf{Project:} \href{https://clausy9.github.io/DemoireEdit/}{\nolinkurl{clausy9.github.io/DemoireEdit}}\\
\textbf{Dataset:} \href{https://huggingface.co/datasets/umu/DailyMoire130}{\nolinkurl{huggingface.co/datasets/umu/DailyMoire130}}}\par\smallskip
\captionsetup{hypcap=false}
\includegraphics[width=\textwidth]{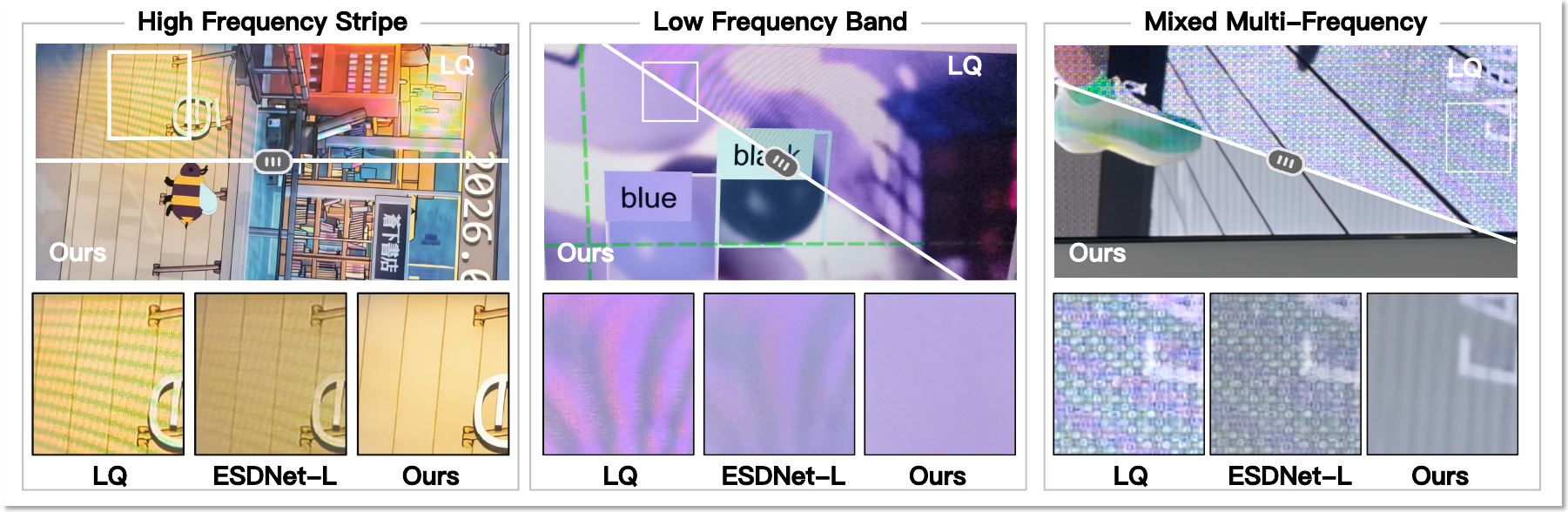}
\captionof{figure}{DailyMoir\'e130 captures challenging moir\'e in everyday screen photography.
Left: high-frequency stripe artifacts.
Middle: low-frequency color banding.
Right: mixed multi-frequency corruption.
Demoir\'eEdit achieves stronger visible moir\'e suppression in these examples.}
\label{fig:teaser}
\par\endgroup
\begin{abstract}
Robust demoir\'eing of screen-recaptured images remains challenging: moir\'e varies in frequency, color, and spatial support across display-camera pairs and capture conditions. Dedicated restoration networks trained on finite paired data struggle with this diversity, especially when low-frequency color banding and high-frequency stripes co-occur. We present Demoir\'eEdit, a content-preserving editing framework that adapts the natural-image prior of a pretrained flow-matching editor to demoir\'eing. First, an endpoint-supervised training objective supplements velocity matching with pixel, perceptual, and frequency-domain restoration losses. Second, a dual-scale condition-shift strategy performs global correction at low resolution, then refines the image at full resolution by shifting the denoising condition from the first-stage output to the original input for detail recovery. We also introduce DailyMoir\'e130, a benchmark of challenging moir\'e in everyday screen photography, comprising 130 smartphone captures across diverse displays with low-frequency color banding, fine high-frequency stripes, and mixed-frequency interference. On the challenging DailyMoir\'e130 benchmark, our method achieves the lowest residual moir\'e area and receives 71.3\% overall-quality preference in a blind Best-of-7 study with 31 participants. We also evaluate on paired FHDMi/UHDM, reducing residual moir\'e confidence by 47.1\%/9.3\% and increasing MUSIQ by 4.3\%/2.0\% over the best baseline for each metric.

\end{abstract}

\section{Introduction}
\begin{wrapfigure}{R}{0.5\textwidth}
  \centering
  \includegraphics[width=\linewidth]{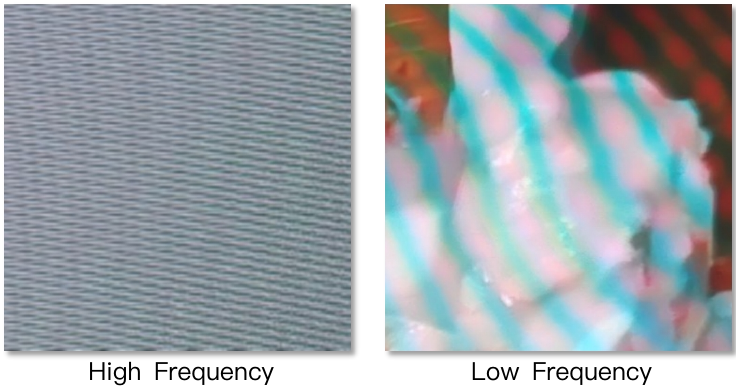}
  \caption{Moir\'e at different frequency scales.}
  \label{fig:high_low}
  \end{wrapfigure}

Moir\'e artifacts arise from the spatial-frequency interference between a display's pixel array and a camera's sensor grid during screen recapture~\citep{sidorov2002suppression,sun2018moire,wang2023coarse}. Unlike noise or blur with a statistically stable spectrum, moir\'e has no fixed frequency signature: low-frequency color banding and high-frequency stripes can co-occur within the same image (Figs.~\ref{fig:teaser} and~\ref{fig:high_low}). Their mixture varies with viewing angle, device pairing, and local scene content~\citep{sun2018moire,yuan2019aim}, making the degradation difficult to characterize with a single spectral model.

Existing approaches train dedicated convolutional or transformer networks from scratch on dataset-specific moir\'e pairs~\citep{sun2018moire,he2019mop,zheng2020image,liu2020wavelet,he2020fhde2net,yu2022uhdm,xiao2024p,zhong2024undem,liu2025freqformer}. These models lack a broader natural-image prior, and their restoration quality can degrade on scenes, devices, or capture conditions outside the training distribution. Moreover, existing benchmarks cover relatively limited device pairings and capture conditions~\citep{sun2018moire,yuan2019aim,yuan2020ntire,he2020fhde2net,yu2022uhdm,xu2024image}. This limited diversity underrepresents complex real-world moir\'e, limiting how accurately these benchmarks assess demoir\'eing performance in the wild.

Our key observation is that dedicated demoir\'eing models trained on limited paired data often struggle to restore complex real-world textures under challenging out-of-distribution moir\'e corruption. Their ability to recover the underlying content is constrained by the visual patterns represented in their training data. In contrast, a pretrained image editor provides a broader natural-image prior acquired through large-scale pretraining, offering the potential to restore content under severe moir\'e corruption beyond the coverage of task-specific paired data~\citep{rombach2022ldm,brooks2023instructpix2pix,saharia2022palette,zhang2023adding,wang2024exploiting,lin2024diffbir,yu2024scaling}. This motivates treating demoir\'eing as a content-preserving editing task and adapting a pretrained flow-matching image editor~\citep{wu2025qwen} to exploit this prior.

However, applying this generative prior to demoir\'eing requires adapting it to the fidelity demands of image restoration. We therefore supplement standard velocity matching with endpoint restoration losses that directly supervise the restored image. The remaining challenge is to handle the co-occurrence of low-frequency color banding and high-frequency stripes while preserving fine details. To address this, Demoir\'eEdit uses a dual-scale condition-shift strategy that first performs global correction at low resolution and then refines the image at full resolution. Since the corrected estimate provides global guidance but lacks fine detail, early denoising steps use the Stage~1 output, whereas later steps return to the original input for detail recovery.

To evaluate challenging moir\'e encountered in daily life, we introduce \textbf{DailyMoir\'e130}: 130 smartphone captures of signage, monitors, and laptops exhibiting low-frequency color banding, fine high-frequency stripes, and mixed-frequency interference (Fig.~\ref{fig:dailymoire_gallery}). These captures lack aligned clean targets; evaluation combines residual moir\'e metrics with separate human judgments of suppression, detail preservation, and overall quality.

On DailyMoir\'e130, our method achieves the lowest residual moir\'e area and receives 71.3\% overall-quality preference in a blind Best-of-7 study with 31 participants. We also evaluate on paired FHDMi/UHDM, reducing residual moir\'e confidence by 47.1\%/9.3\% and increasing MUSIQ by 4.3\%/2.0\% over the best baseline for each metric (Tab.~\ref{tab:main_paired}). Our contributions are:
\begin{itemize}[leftmargin=*]
\item We reformulate demoir\'eing as content-preserving editing and introduce endpoint-supervised flow training to adapt a pretrained editor for faithful restoration.
\item We propose dual-scale condition-shift inference to separate global moir\'e correction from full-resolution detail recovery.
\item We introduce \textbf{DailyMoir\'e130}, a benchmark targeting challenging moir\'e in everyday screen photography, with a reference-free evaluation protocol. The dataset is publicly available to support future comparisons (Appendix~\ref{app:daily}).
\end{itemize}

\section{Related Work}

\paragraph{Image demoir\'eing.}
Early demoir\'eing methods relied on hand-crafted spectral priors or signal-processing
assumptions to suppress periodic moir\'e patterns~\citep{sidorov2002suppression,
siddiqui2009hardware,liu2015moire}. With the rise of deep learning, Sun et al.~\citep{sun2018moire} introduced the first
CNN-based solution, followed by frequency-aware designs using learnable bandpass
filters~\citep{zheng2020image}, wavelet decomposition~\citep{liu2020wavelet,luo2020deep,
yeh2024multibranch}, and multiscale fusion~\citep{peng2024image}.
Subsequent work extends demoir\'eing to ultra-high-definition inputs~\citep{yu2022uhdm,xiao2024p}, unpaired data~\citep{zhong2024undem}, RAW-domain processing~\citep{xu2024image,yang2025dsdnet},
and universal data generation~\citep{yang2025unidemoire,choi2024screensim}. In particular,
Freqformer~\citep{liu2025freqformer} explicitly separates moir\'e artifacts into high-frequency
localized textures and low-frequency color distortions. While these approaches emphasize task-specific frequency modeling, we adapt a pretrained editor's natural-image prior to challenging real-world moir\'e.
\paragraph{Generative priors for image restoration and editing.}
Latent diffusion models~\citep{rombach2022ldm} provide
a powerful visual prior for high-quality generation, and instruction-based or condition-based
editing frameworks such as SDEdit~\citep{meng2021sdedit}, InstructPix2Pix~\citep{brooks2023instructpix2pix},
Palette~\citep{saharia2022palette}, ControlNet~\citep{zhang2023adding}, and Qwen-Image
Edit~\citep{wu2025qwen} enable image manipulation under text or visual conditions.
In low-level vision, pretrained diffusion priors have been adapted to real-world super-resolution
and blind restoration~\citep{wang2024exploiting,lin2024diffbir,yu2024scaling,wu2024one,wu2024seesr},
while prompt- or language-guided restoration methods learn degradation-adaptive controls for
all-in-one restoration~\citep{potlapalli2023promptir,luo2023controlling,cui2025adair,
zamfir2025complexity}. However, these methods do not directly address mixed-frequency moir\'e, which motivates our task-specific adaptation.

\paragraph{Flow matching and trajectory supervision.}
Flow matching trains continuous normalizing flows by regressing vector fields along prescribed
probability paths~\citep{lipman2023flowmatching}, and rectified flow learns straight transport
paths that enable efficient ODE sampling~\citep{liu2022flow}. These objectives are
effective for generative modeling, but do not by themselves guarantee pixel-faithful restoration
endpoints. Related acceleration and consistency objectives supervise shortened or distilled
trajectories~\citep{salimans2022progressive,song2023consistency}. We therefore complement velocity supervision with pixel, perceptual, and frequency losses~\citep{zhang2018lpips,jiang2021ffl}.

\section{Method}
\begin{figure}[!ht]
  \centering
  \includegraphics[width=\textwidth]{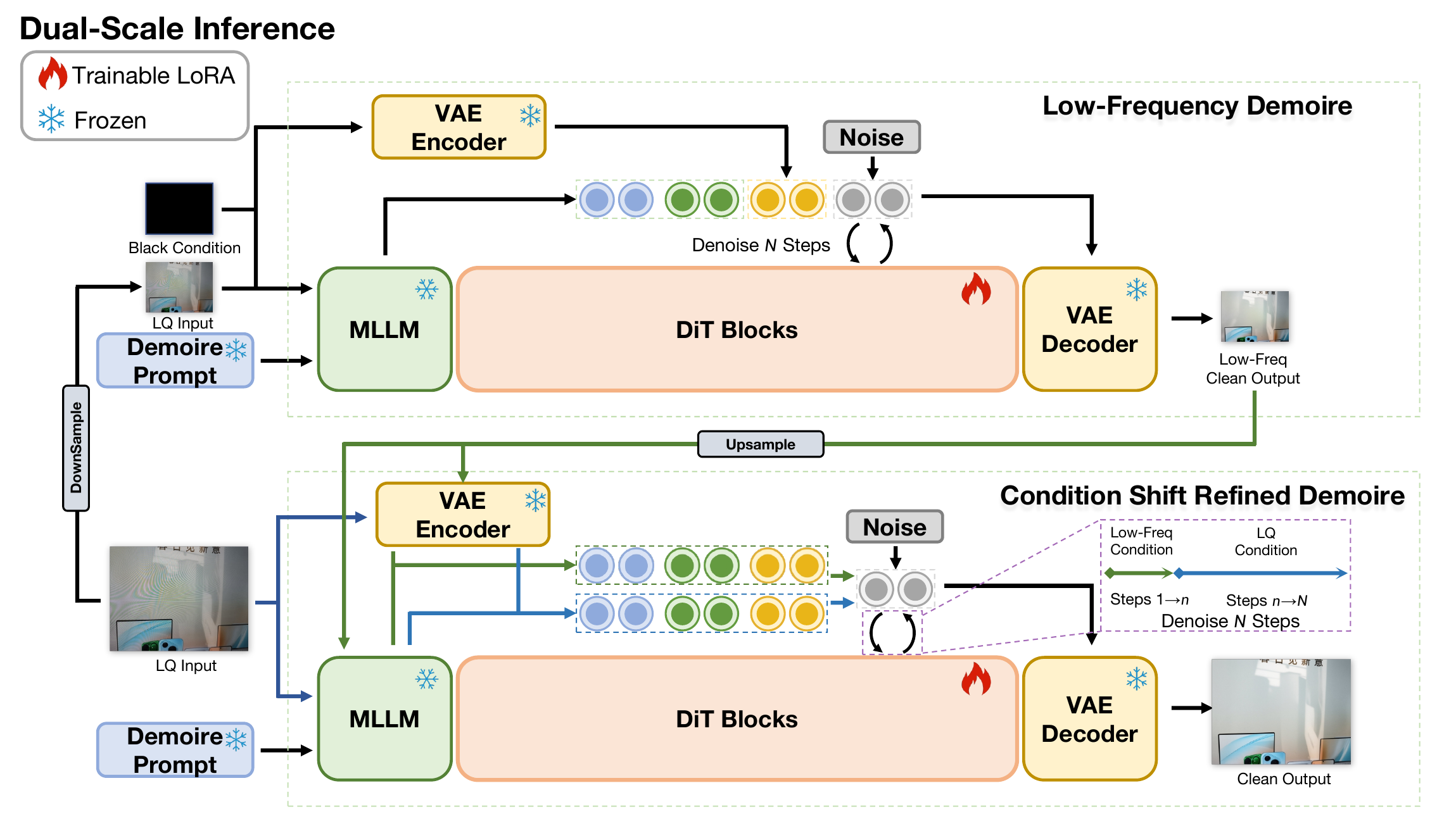}
  \caption{Overview of our dual-scale inference. \textbf{Stage 1, Low-frequency Demoir\'e:} the low-quality (LQ) moir\'e image is downsampled and paired with a black auxiliary reference as the condition. \textbf{Stage 2, Condition Shift Refined Demoir\'e:} the low-frequency output of Stage 1 is used as a condition for the first $n$ steps for structured guidance. The condition is shifted back to the original LQ image for the remaining $N{-}n$ steps for high-frequency content and detail recovery.}
  \label{fig:framework}
\end{figure}

Our model builds on Qwen-Edit~\citep{wu2025qwen}, a flow-matching image editor conditioned on a source image, edit references, and a text prompt. We first define the restoration objective and training procedure, then describe dual-scale inference (Fig.~\ref{fig:framework}) and the training data.

\subsection{Problem Definition}\label{sec:problem}
Given a moir\'e-corrupted image $I_m \in \mathbb{R}^{H \times W \times 3}$ and its clean counterpart $I_c$, we learn a mapping $\hat{I} = \mathcal{F}(I_m)$ that removes moir\'e while preserving scene content, such that $\hat{I} \approx I_c$.

\subsection{Endpoint-supervised flow training}\label{sec:training}
Let $v_{\theta}$ denote the pretrained editor's velocity network, and $\mathcal{E}$ and $\mathcal{D}$ its VAE encoder and decoder. For clean latent $z_0 = \mathcal{E}(I_c)$ and noise $z_1 \sim \mathcal{N}(0,\mathbf{I})$, the flow interpolation is $z_t = (1-t)\cdot z_0 + t\cdot z_1$, with $t\in[0,1]$ and target velocity $v^{\star}=z_0-z_1$. Conditioning $\xi=(I_m,e,p_{\mathrm{dm}})$ comprises the input, an edit reference, and a fixed task prompt shared by training and inference (Appendix~\ref{app:recipe}). Standard flow matching supervises single-timestep velocities~\citep{lipman2023flowmatching,liu2022flow}:
\begin{equation}
\mathcal{L}_{\mathrm{FM}} = \mathbb{E}_{t, z_0, z_1} \left[ \| v_{\theta}(z_t, t, \xi) - v^{\star} \|_2^2 \right],
\end{equation}
To supervise restoration fidelity at the image endpoint, we unroll $K{=}5$ steps and retain velocity supervision along the trajectory. With a descending schedule $t^{(0)}>\cdots>t^{(K)}$ and $\hat{z}^{(0)}=z_{t^{(0)}}$, the updates are
\begin{equation}
\hat{z}^{(k)} = \mathrm{Step}\!\left(\hat{z}^{(k-1)},\; v_{\theta}(\hat{z}^{(k-1)},\, t^{(k)},\, \xi)\right), \qquad k = 1,\ldots,K.
\label{eq:unroll}
\end{equation}
The velocity loss over the unrolled trajectory is
\begin{equation}
\mathcal{L}_{\mathrm{vel}} = \frac{1}{K}\sum_{k=1}^{K} \| v_{\theta}(\hat{z}^{(k-1)},\, t^{(k)},\, \xi) - v^{\star} \|_2^2.
\label{eq:lvel}
\end{equation}

We decode the endpoint $\hat{I}=\mathcal{D}(\hat{z}^{(K)})$ and optimize
\begin{equation}
\mathcal{L}_{\mathrm{total}} = \mathcal{L}_{\mathrm{vel}} + \mathcal{L}_{\mathrm{restore}},
\label{eq:ltotal}
\end{equation}
where $\mathcal{L}_{\mathrm{restore}}$ combines pixel-space $\ell_1$, LPIPS~\citep{zhang2018lpips}, and focal frequency loss~\citep{jiang2021ffl}, each referenced to the clean target $I_c$. Full definitions and weights are in Appendix~\ref{app:loss}.

\subsection{Dual-scale frequency decoupling}\label{sec:dualscale}
We conceptually decompose the corruption into low-frequency color bands and high-frequency stripes:
\begin{equation}
I_m = I_c + M_{\mathrm{low}} + M_{\mathrm{high}},
\label{eq:moire_decomp}
\end{equation}
where $M_{\mathrm{low}}$ and $M_{\mathrm{high}}$ denote low- and high-frequency moir\'e components.

\paragraph{Stage 1: prior-guided low-frequency correction.}
Let $\downarrow_s$ and $\uparrow_s$ denote downsampling and upsampling by factor $s$, and $f_{\mathrm{Ny}}$ the original Nyquist frequency. Downsampling reduces the Nyquist limit to $f_{\mathrm{Ny}}/s$, folding higher-frequency moir\'e components into the lower band through aliasing:
\begin{equation}
\downarrow_s(I_m) = \downarrow_s(I_c) + \underbrace{\downarrow_s(M_{\leq f_{\mathrm{Ny}}/s}) + \mathcal{A}_s(M_{> f_{\mathrm{Ny}}/s})}_{\displaystyle\tilde{M},\;\; \mathrm{supp}(\hat{\tilde{M}}) \,\subseteq\, [0,\, f_{\mathrm{Ny}}/s]},
\label{eq:alias}
\end{equation}
where $\mathcal{A}_s$ is the aliasing operator and $M=M_{\mathrm{low}}+M_{\mathrm{high}}$. The resulting lower-band representation makes Stage~1 focus on global structure and low-frequency appearance.

We pair the downsampled input with a black auxiliary reference:
\begin{equation}
I_{lf} = f_{\theta}(\downarrow_s(I_m), e=[\downarrow_s(I_m),\, \mathbf{0}], p_{\mathrm{dm}}),
\label{eq:stage1}
\end{equation}
where $f_{\theta}$ denotes denoising from noise followed by VAE decoding. The references are VAE-encoded for conditioning; the black reference is intended to reduce copying of corrupted color and encourage use of the clean-image prior. Tab.~\ref{tab:ablation_component} evaluates this choice.

\paragraph{Stage 2: condition-shift refinement at full resolution.}

We align conditioning with denoising progression: early high-noise steps use the Stage~1 result for structure and color, while later low-noise steps use the original input for texture and detail. The two edit latents at the target resolution are
\begin{equation}
e_{\mathrm{mid}} = \mathcal{E}\!\left(\uparrow_s(I_{lf})\right), \qquad
e_{\mathrm{lq}} = \mathcal{E}(I_m),
\label{eq:edit_latents}
\end{equation}
For $N$ steps in sampling order (high to low noise) and switching ratio $r\in[0,1]$, the condition is
\begin{equation}
e^{(j)} =
\begin{cases}
e_{\mathrm{mid}}, & 1 \le j \le n,\\
e_{\mathrm{lq}}, & n < j \le N,
\end{cases}
\qquad n = \lceil r N \rceil.
\label{eq:edit_schedule}
\end{equation}
The denoising trajectory starts from $z^{(0)}\!\sim\!\mathcal{N}(0,\mathbf{I})$
and updates
\begin{equation}
z^{(j)} = \mathrm{Step}\!\left(z^{(j-1)},\; v_{\theta}(z^{(j-1)},\, t^{(j)},\, e^{(j)},\, p_{\mathrm{dm}})\right),\qquad j=1,\ldots,N,
\label{eq:stage2_step}
\end{equation}
yielding the restored image $\hat{I} = \mathcal{D}(z^{(N)})$.
We use $r=0.3$ to balance artifact removal and detail preservation; Sec.~\ref{sec:ablations} examines this trade-off.

\subsection{Mixed demoir\'e data}\label{sec:data}
We train on a mixed corpus $\mathcal{D} = \mathcal{D}_{\mathrm{public}} \cup \mathcal{D}_{\mathrm{doc}} \cup \mathcal{D}_{\mathrm{self-rec}}$. $\mathcal{D}_{\mathrm{public}}$ combines paired benchmarks~\citep{sun2018moire,yuan2019aim,yuan2020ntire,he2020fhde2net,yu2022uhdm,xu2024image}, the UniDemoire augmentation pipeline~\citep{yang2025unidemoire}, and the screen-photo simulator of~\citet{choi2024screensim}. $\mathcal{D}_{\mathrm{doc}}$ contains high-contrast documents and forms from DocVQA~\citep{mathew2021docvqa} and XFUND~\citep{xu2022xfund} to strengthen text preservation. $\mathcal{D}_{\mathrm{self-rec}}$ uses identical clean images as input and target to preserve artifact-free content and discourage over-editing.

\section{Experiments}
\subsection{Experimental Setup}\label{sec:setup}

\paragraph{Implementation details.}
We fine-tune the pretrained Qwen-Edit~\citep{wu2025qwen} flow-matching editor via LoRA~\citep{hu2022lora} adapters on DiT using our endpoint-supervised objective (Sec.~\ref{sec:training}) on the mixed datasets described in Sec.~\ref{sec:data}, which aggregates six public demoir\'eing datasets. We use a batch size of 96 and a patch size of $256\times256$ with a cosine learning-rate schedule decaying from $1\times10^{-4}$ to $1\times10^{-7}$. We fine-tune for 3k optimizer steps. All experiments are conducted on 4 NVIDIA H100 GPUs. During inference, Stage 1 resizes the input to a short side of 1024 pixels and uses 10 denoising steps. Stage 2 starts from latent noise at full resolution and uses $N{=}10$ steps with switching ratio $r{=}0.3$. The fixed prompt and full training recipe are given in Appendix~\ref{app:recipe}.

\begin{figure}[!t]
  \centering
  \includegraphics[width=\textwidth]{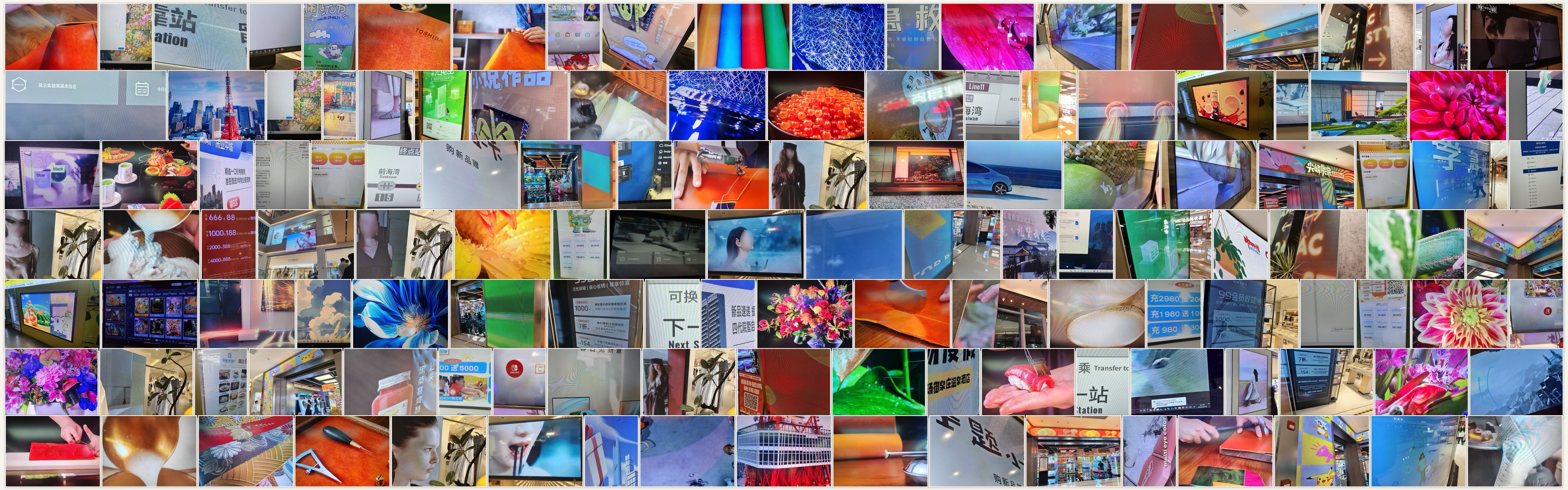}
  \caption{DailyMoir\'e130 captures challenging moir\'e on everyday displays (signage, monitors, laptops) using multiple smartphones. \textbf{The dataset is publicly available;} evaluation details are provided in Appendix~\ref{app:daily}.}
  \label{fig:dailymoire_gallery}
\end{figure}

\paragraph{Benchmarks and metrics.}
To assess restoration in challenging everyday screen-capture scenarios, we collect \textbf{DailyMoir\'e130}, a benchmark of 130 screen photographs from shopping-mall signage, desktop monitors, and laptops. These captures exhibit low-frequency color banding, fine high-frequency stripes, and mixed-frequency interference, and have no aligned clean targets. They were captured with iPhone~17, vivo~X200s, Honor 90 Pro, Huawei Enjoy 90 Pro Max, and iPhone 16 Pro Max (Fig.~\ref{fig:dailymoire_gallery}). We also use FHDMi~\citep{he2020fhde2net} with 2019 1080p images and UHDM~\citep{yu2022uhdm} with 500 4K images for paired evaluation. Due to the spatially non-uniform nature of moir\'e, full-image fidelity metrics such as PSNR and SSIM~\citep{wang2004image} can be dominated by artifact-free regions and fail to reflect actual moir\'e removal quality.

We therefore additionally report two auxiliary metrics based on a pretrained moir\'e detector $\mathcal{D}$~\citep{yang2023doing} called Moir\'eDet. Given a restored image $\hat{I}$, we divide it into $P$ overlapping patches and obtain a moir\'e probability $c_i \in [0,1]$ for each patch, where a higher $c_i$ indicates stronger moir\'e presence. Let $M$ denote the number of patches whose moir\'e probability exceeds $0.05$. We define:
\begin{equation}
\mathrm{RMC} = \frac{1}{P}\sum_{i=1}^{P} c_i, \qquad
\mathrm{RMA} = \frac{M}{P}.
\label{eq:moire_metrics}
\end{equation}
RMC measures the average moir\'e confidence across all patches, and RMA captures the proportion of patches with detectable residual artifacts. We also report four no-reference image quality metrics: MUSIQ~\citep{ke2021musiq}, MANIQA~\citep{yang2022maniqa}, CLIP-IQA+~\citep{wang2022exploring}, and TOPIQ~\citep{chen2024topiq} for paired datasets and CLIP-IQA+ for DailyMoir\'e130, which assess perceptual quality without requiring ground-truth references.

\paragraph{Compared methods.}
We compare with ESDNet-L~\citep{yu2022uhdm}, a dedicated demoir\'eing network, along with three general-purpose image restoration methods: MoCE-IR~\citep{zamfir2025complexity}, AdaIR~\citep{cui2025adair}, and the diffusion-based OSEDiff~\citep{wu2024one}.
These four baselines are retrained on the same mixed training data for 50k optimizer
updates, using their official training configurations.
We additionally compare with MZNet~\citep{mznet2025}, trained on the same mixed data for 50k updates, and a UniDemoir\'e-style augmentation control~\citep{yang2025unidemoire}, obtained by fine-tuning ESDNet-L for 10k updates with real paired and synthetic data. Freqformer code and weights were unavailable during these experiments.

\begin{table}[!htb]
\centering
\caption{Evaluation on DailyMoir\'e130. Blind Best-of-7 preference (\%) from 31 participants, with 18 cases each. Uncertain responses remain in the denominator. The input is shown only as a reference.}
\label{tab:realdata}
\papertablestyle
\begin{tabular*}{\linewidth}{@{\extracolsep{\fill}}lrrrrrr@{}}
\toprule
& \multicolumn{3}{c}{User preference $\uparrow$} & \multicolumn{3}{c}{Automatic metrics}\\
\cmidrule(lr){2-4}\cmidrule(l){5-7}
Method & Moir\'e & Detail & Overall & CLIP-IQA+$\uparrow$ & RMC$\downarrow$ & RMA$\downarrow$\\
\midrule
Input & --- & --- & --- & .5690 & .0523 & .2507\\
ESDNet-L & 8.6 & 8.6 & 8.8 & .5478 & .0340 & .1599\\
MoCE-IR & 1.8 & 2.3 & 2.2 & .5277 & .0341 & .1602\\
AdaIR & 1.4 & 2.5 & 2.0 & .5464 & .0447 & .2163\\
OSEDiff & 2.3 & 3.4 & 2.7 & .5671 & .0163 & .0794\\
MZNet & 2.7 & 2.9 & 2.5 & .5287 & .0334 & .1562\\
Uni-control & 5.4 & 6.8 & 6.6 & .5296 & \textbf{.0113} & .0567\\
\midrule
\textbf{Ours} & \textbf{74.7} & \textbf{68.5} & \textbf{71.3} & \textbf{.5928} & .0115 & \textbf{.0551}\\
\bottomrule
\end{tabular*}
\end{table}
\subsection{Quantitative Evaluation}
Tab.~\ref{tab:realdata} evaluates all methods on our proposed DailyMoir\'e130, which targets challenging moir\'e in everyday screen photography. In a blind Best-of-7 study with 31 participants, our method is preferred in 74.7\% of trials for moir\'e suppression, 68.5\% for detail preservation, and 71.3\% for overall quality. Participants assess detail preservation separately against the visible input using full-image views and magnified crops (Appendix~\ref{app:studies}). The strongest baseline, ESDNet-L, receives 8.8\% overall preference. The no-reference perceptual metric further supports this finding, with our method achieving the highest CLIP-IQA+ (0.5928). It also achieves the lowest residual moir\'e area (0.0551), while UniDemoir\'e-control has slightly lower RMC. These results demonstrate effective moir\'e suppression and strong human preference on challenging real-world captures.

Each participant completes 18 Best-of-7 trials, collectively covering all 130 images, with method identities hidden and candidate positions randomized and counterbalanced. Uncertain responses remain in the denominator. The overall-quality preference rate has a participant-clustered 95\% confidence interval of 65.1--76.9\%, and the result remains consistent under stricter attention checks (Appendix~\ref{app:studies}).

We also evaluate on the paired benchmarks FHDMi and UHDM (Tab.~\ref{tab:main_paired}). Our method ranks first on MUSIQ, CLIP-IQA+, and TOPIQ, with the lowest detector-based RMC on both datasets.

\begin{table}[!htb]
\centering
\caption{Paired-benchmark comparison. Best in \textbf{bold}, second-best \underline{underlined}. TOPIQ is no-reference. Uni-control denotes UniDemoir\'e-style augmentation of ESDNet-L. Remaining metrics are in Appendix~\ref{app:paired-metrics}.}
\label{tab:main_paired}\label{tab:original-paired}
\papertablestyle
\setlength{\tabcolsep}{1.5pt}
\begin{tabular*}{\linewidth}{@{\extracolsep{\fill}}lrrrrrrrrrr@{}}
\toprule
& \multicolumn{5}{c}{FHDMi} & \multicolumn{5}{c}{UHDM}\\
\cmidrule(lr){2-6}\cmidrule(l){7-11}
Method & PSNR$\uparrow$ & MUSIQ$\uparrow$ & \shortstack{CLIP-\\IQA+$\uparrow$} & TOPIQ$\uparrow$ & RMC$\downarrow$ & PSNR$\uparrow$ & MUSIQ$\uparrow$ & \shortstack{CLIP-\\IQA+$\uparrow$} & TOPIQ$\uparrow$ & RMC$\downarrow$\\
\midrule
Input & 18.00 & 56.76 & .528 & .442 & .0395 & 17.12 & 64.04 & .617 & .537 & .0652\\
ESDNet-L & 20.67 & 55.56 & .523 & .408 & .0175 & 21.16 & 65.44 & .625 & .556 & .0139\\
MoCE-IR & \underline{22.40} & 56.33 & .521 & .424 & .0150 & 20.92 & 64.23 & .600 & .521 & .0189\\
AdaIR & 21.80 & 56.00 & .508 & .425 & .0205 & 20.61 & 63.27 & .587 & .509 & .0342\\
OSEDiff & 18.04 & \underline{59.00} & \underline{.545} & \underline{.476} & \underline{.0121} & 17.57 & 65.16 & .612 & .555 & \underline{.0043}\\
MZNet & \textbf{23.92} & 56.95 & .542 & .437 & .0183 & \textbf{21.67} & \underline{66.10} & \underline{.635} & \underline{.566} & .0149\\
Uni-control & 20.92 & 55.61 & .531 & .412 & .0144 & \underline{21.21} & 65.51 & .630 & .564 & .0110\\
\midrule
\textbf{Ours} & 21.09 & \textbf{61.55} & \textbf{.585} & \textbf{.508} & \textbf{.0064} & 19.99 & \textbf{67.43} & \textbf{.656} & \textbf{.585} & \textbf{.0039}\\
\bottomrule
\end{tabular*}
\end{table}

Our method reduces RMC by 47.1\% and 9.3\%, respectively, relative to the strongest baseline for this metric, and improves MUSIQ by 4.3\% and 2.0\%. While regression-based methods achieve higher PSNR, they retain more residual moir\'e, indicating a trade-off between pixel fidelity and artifact removal. Appendix~\ref{app:paired-metrics} reports SSIM, LPIPS, and MANIQA for all methods.

\paragraph{Content and color fidelity.}
Moreover, we assess content fidelity against clean references on UHDM to complement the dedicated human detail assessment on DailyMoir\'e130. Our method achieves the highest TOPIQ-FR of 0.670, compared with 0.664 for the strongest baseline, MZNet. It also obtains the second-best PieAPP of 0.984 and AHIQ of 0.435 among the compared methods, while its chroma and hue errors remain comparable to those of feed-forward restorers (Appendix~\ref{app:paired-metrics}, Tab.~\ref{tab:content}). Unlike TOPIQ in Tab.~\ref{tab:main_paired}, TOPIQ-FR uses the clean reference. These results provide complementary evidence of content preservation alongside the gains in moir\'e suppression.

\subsection{Qualitative Evaluation}
\begin{figure}[!htb]
\centering
\setlength{\tabcolsep}{1pt}
\renewcommand{\arraystretch}{0.6}
\begin{tabular}{ccccccc}
\includegraphics[width=0.138\textwidth]{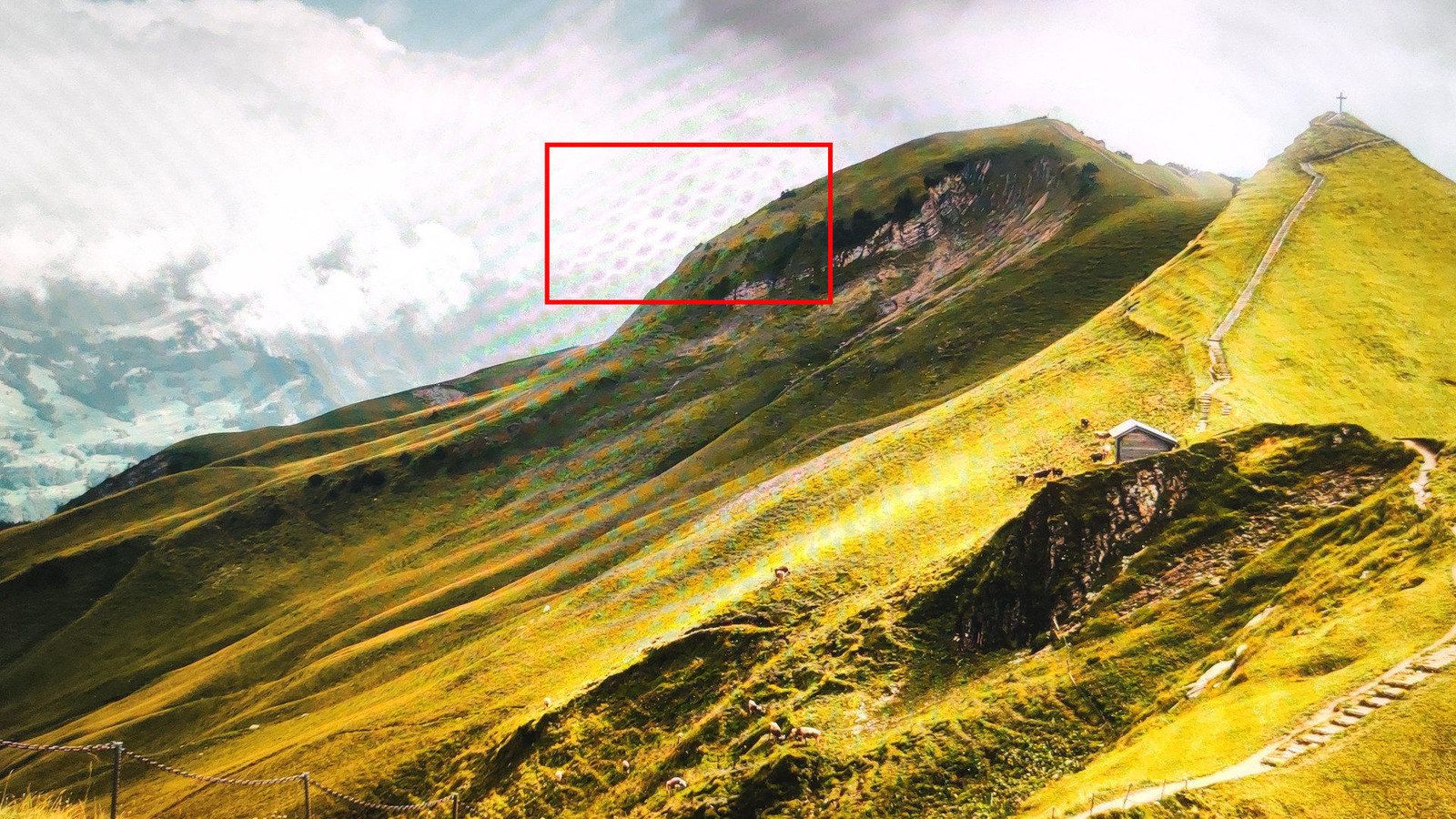} &
\includegraphics[width=0.138\textwidth]{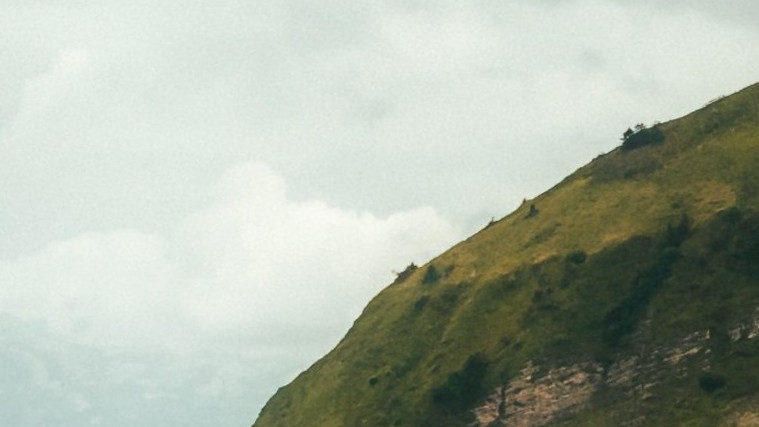} &
\includegraphics[width=0.138\textwidth]{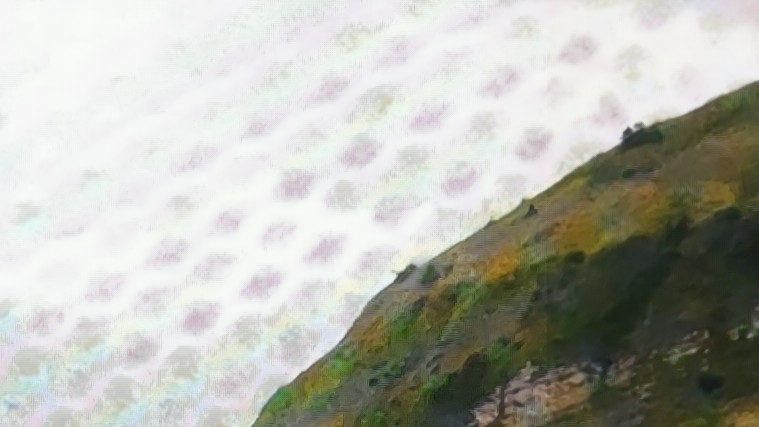} &
\includegraphics[width=0.138\textwidth]{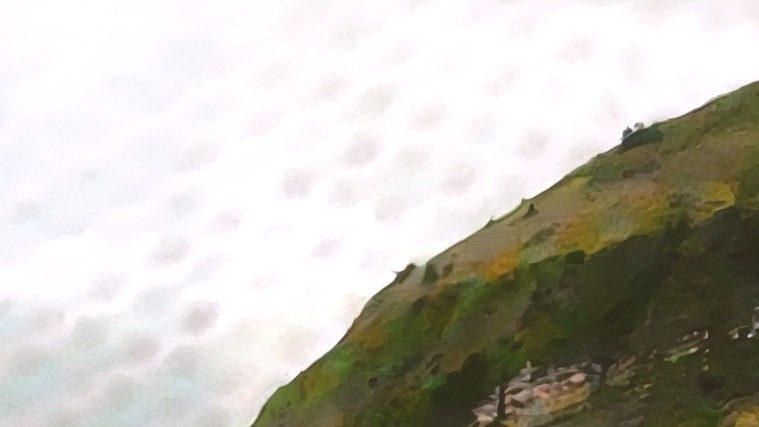} &
\includegraphics[width=0.138\textwidth]{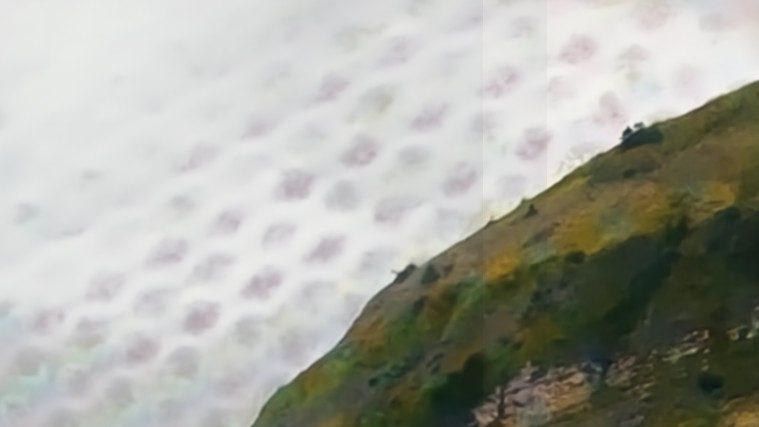} &
\includegraphics[width=0.138\textwidth]{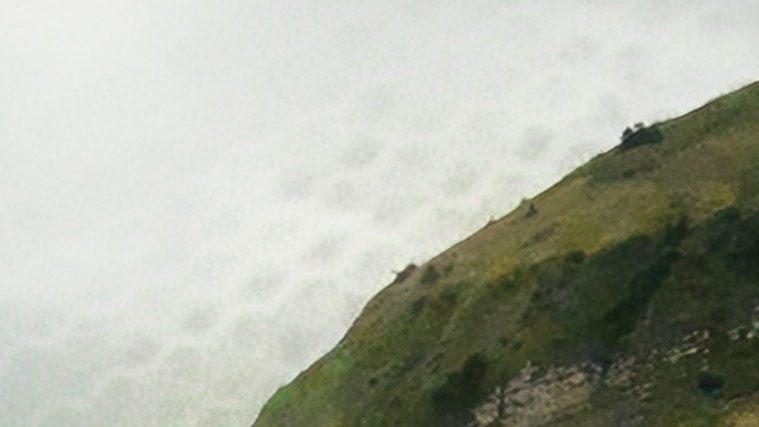} &
\includegraphics[width=0.138\textwidth]{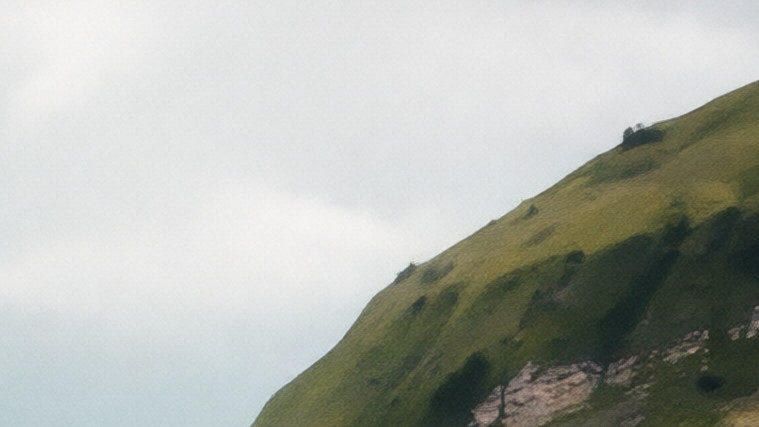} \\
\includegraphics[width=0.138\textwidth]{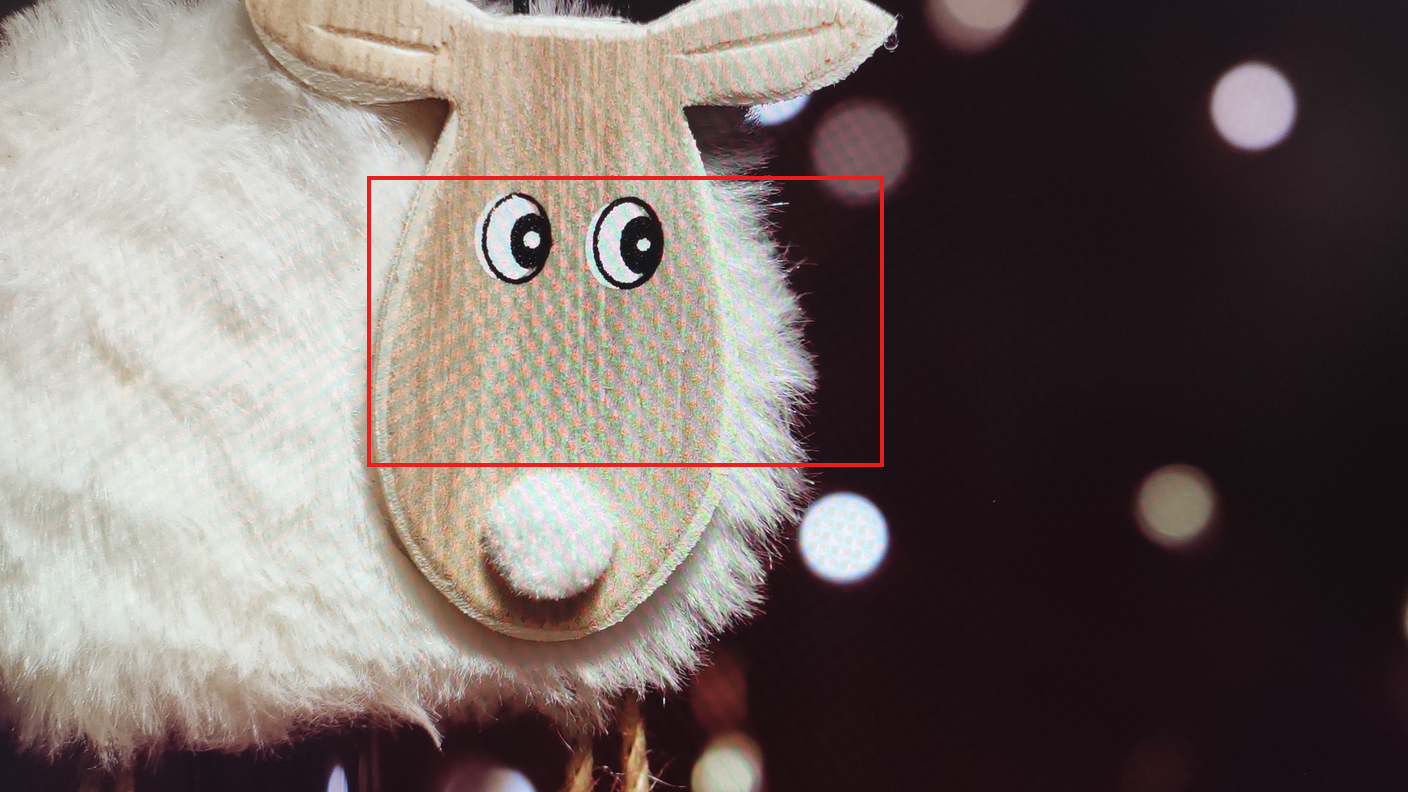} &
\includegraphics[width=0.138\textwidth]{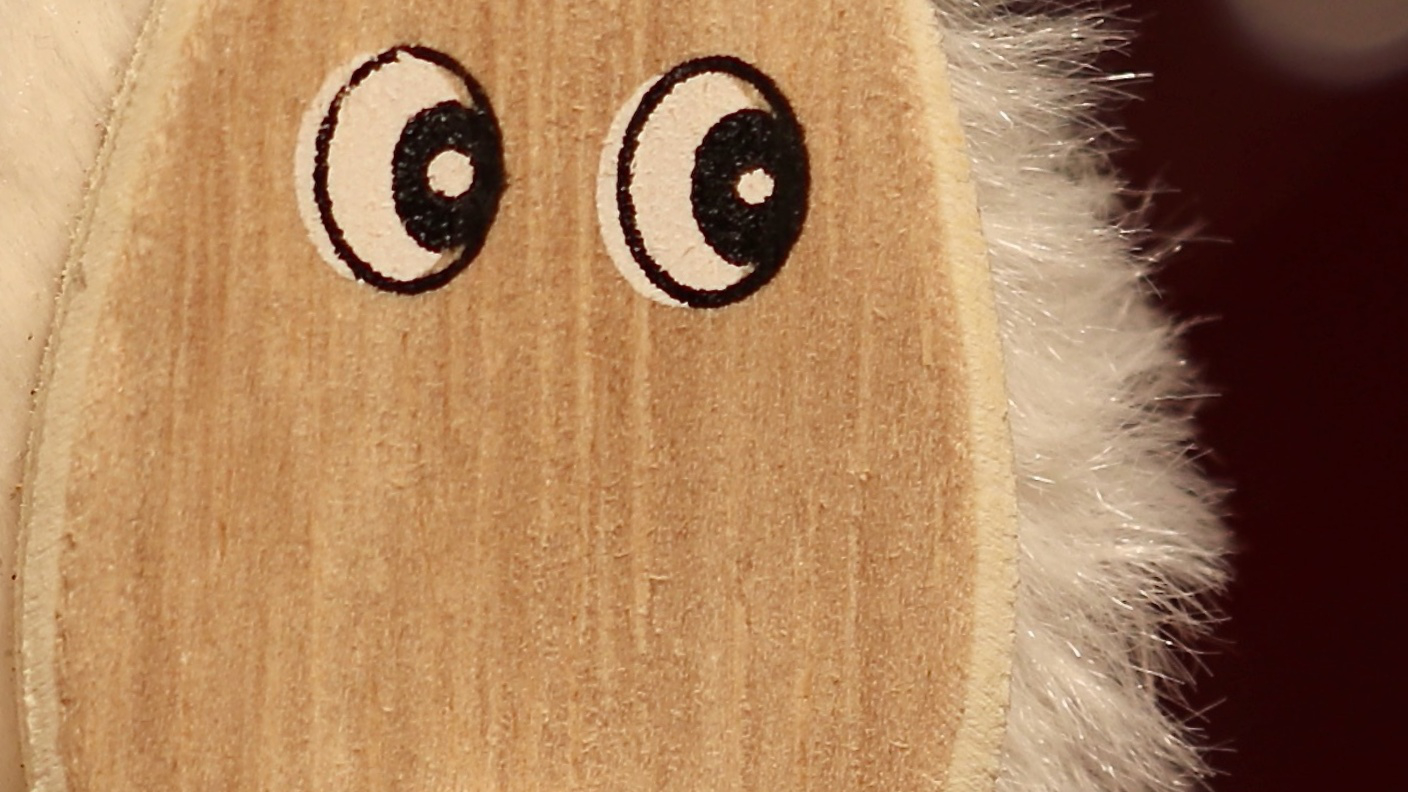} &
\includegraphics[width=0.138\textwidth]{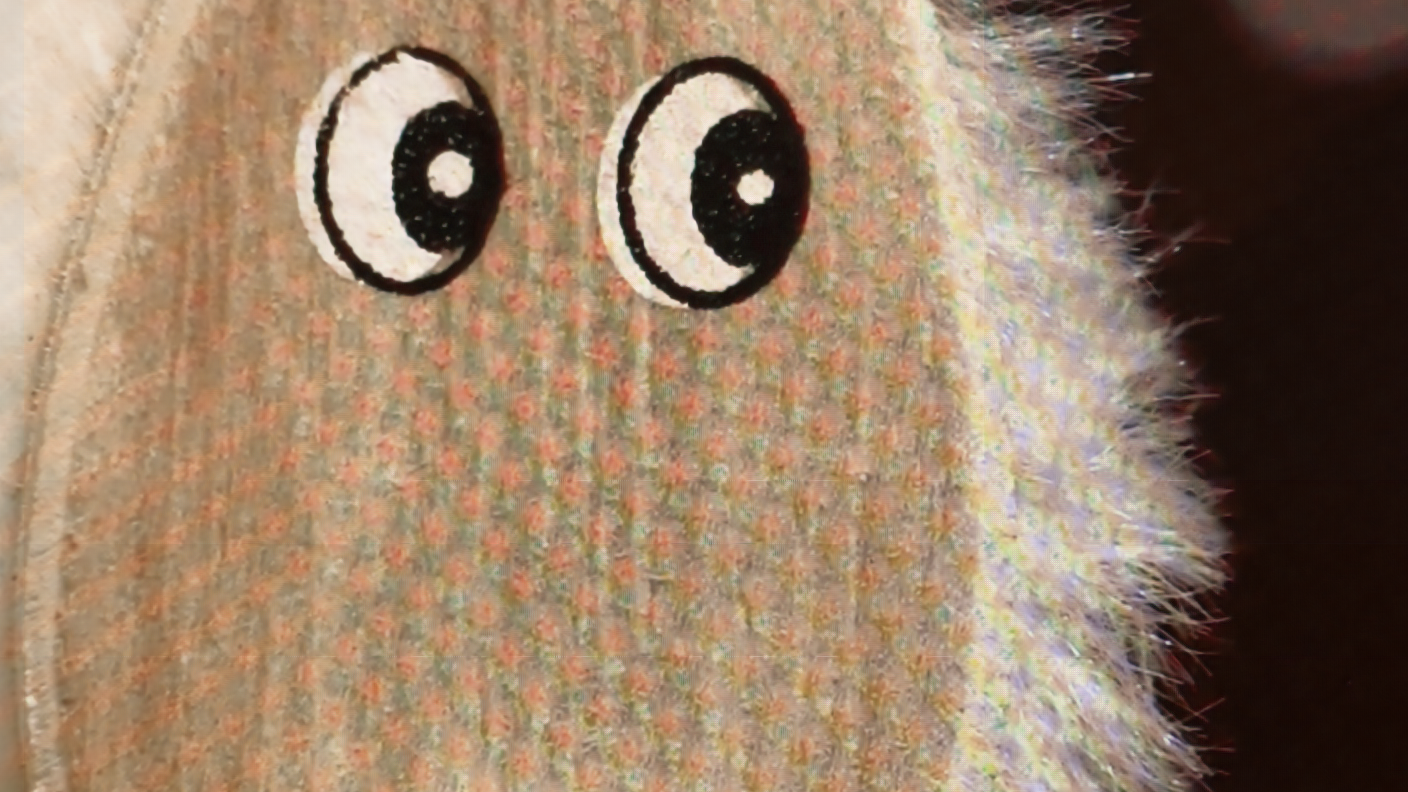} &
\includegraphics[width=0.138\textwidth]{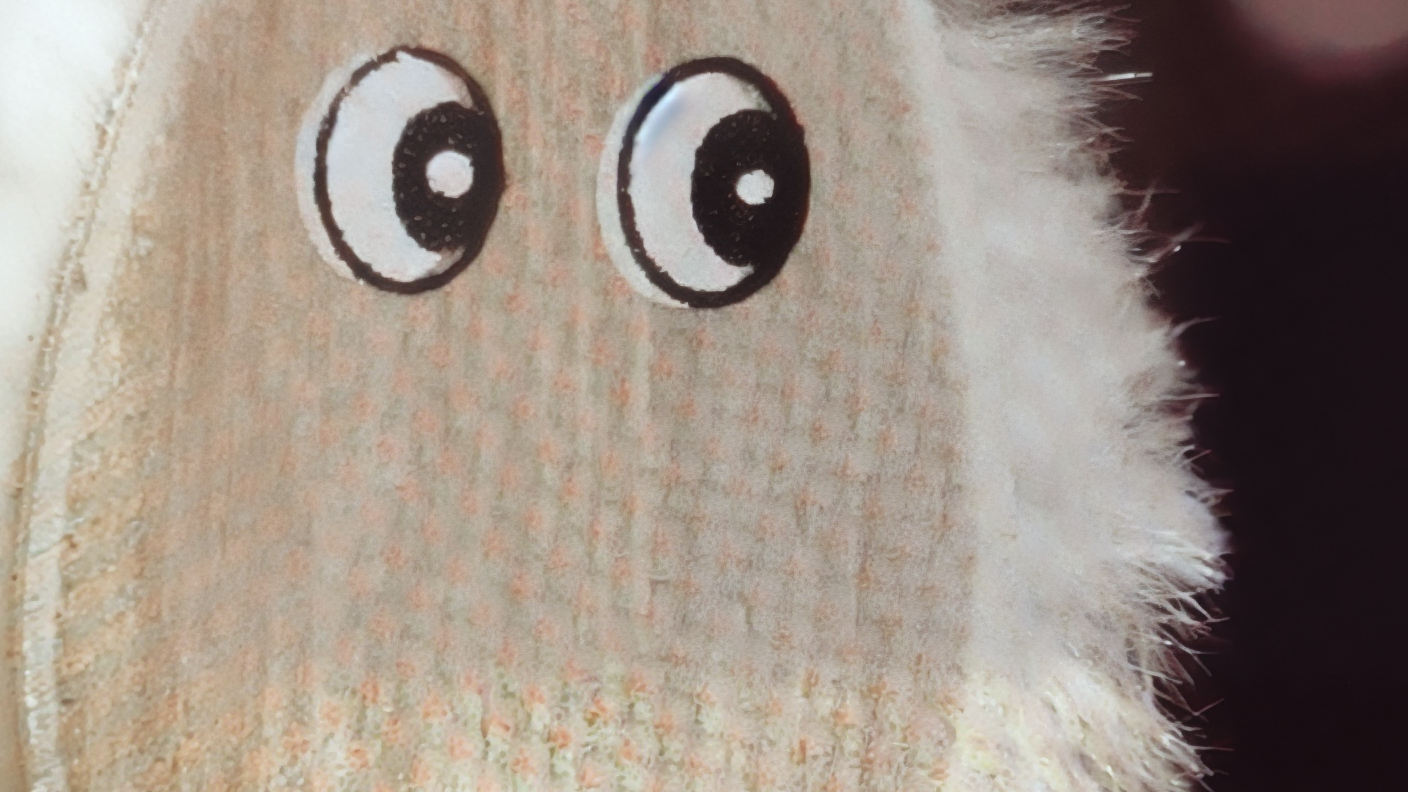} &
\includegraphics[width=0.138\textwidth]{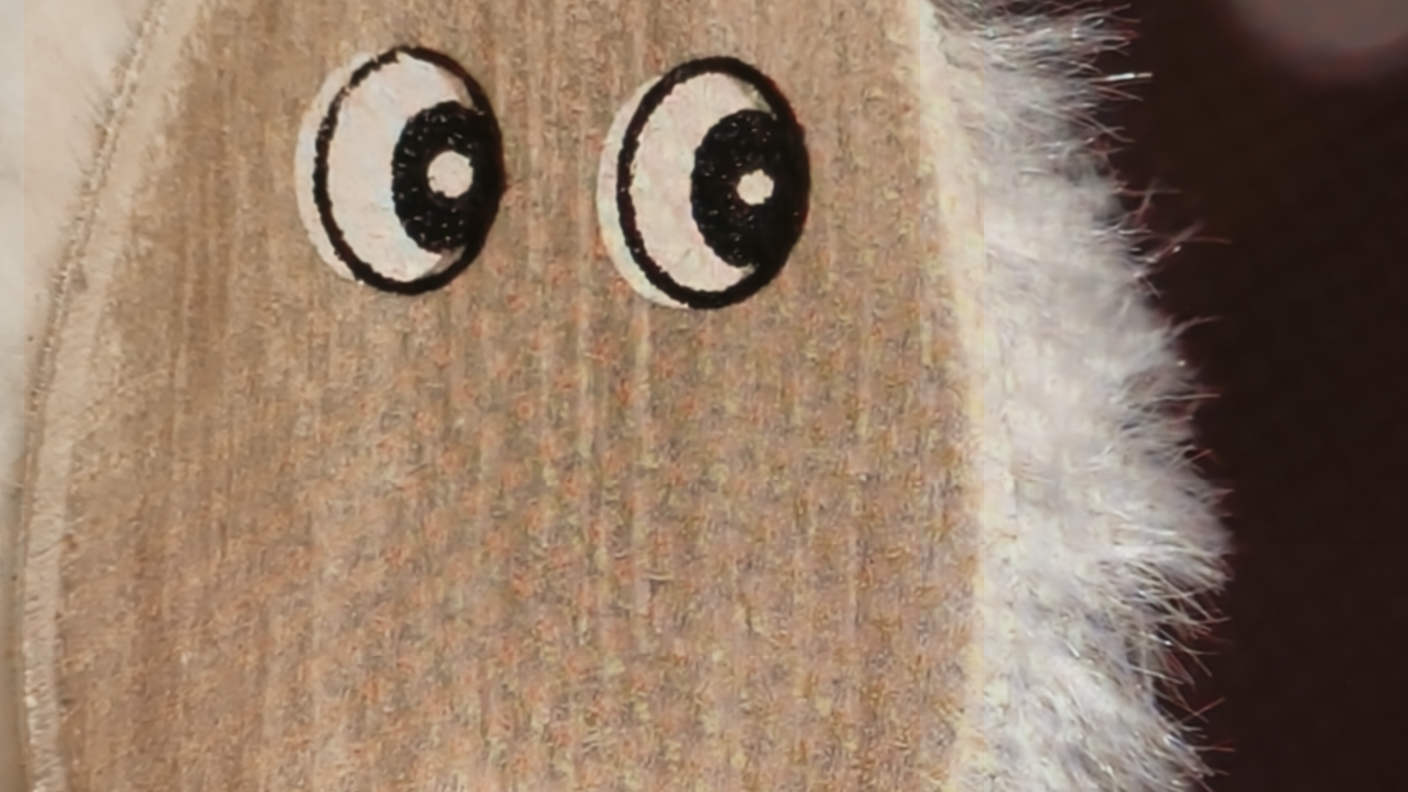} &
\includegraphics[width=0.138\textwidth]{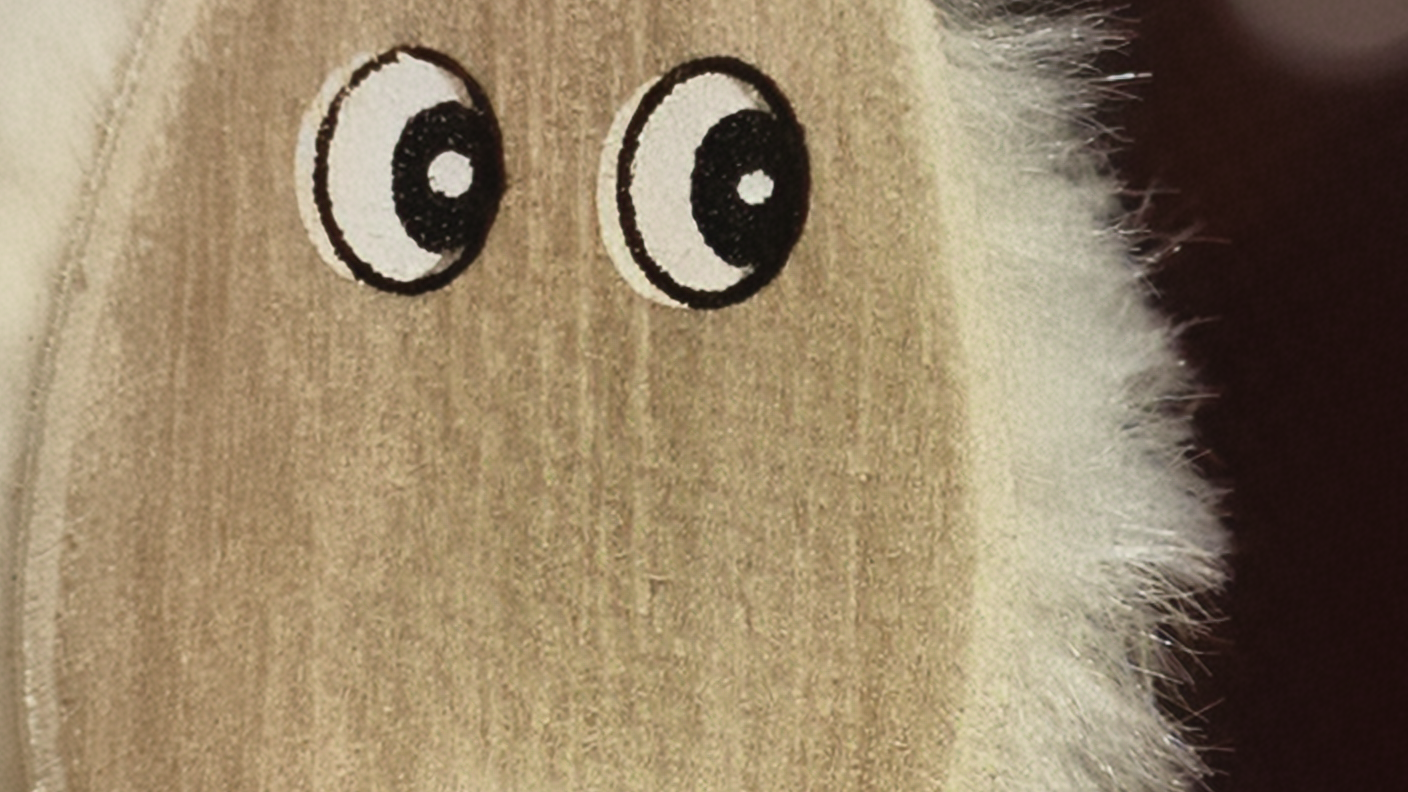} &
\includegraphics[width=0.138\textwidth]{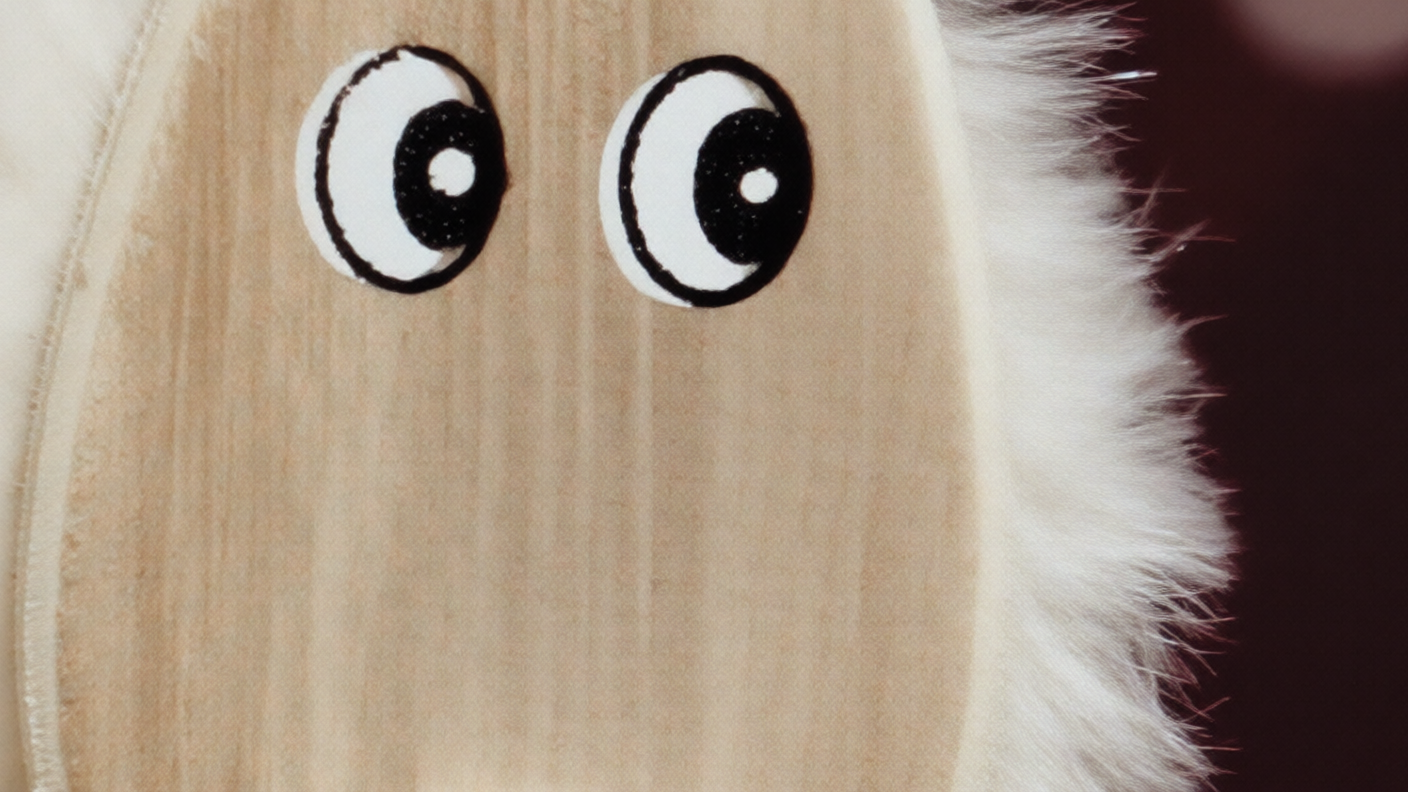} \\[2pt]
{\scriptsize Input} &
{\scriptsize GT} &
{\scriptsize AdaIR} &
{\scriptsize OSEDiff} &
{\scriptsize MoCE-IR} &
{\scriptsize ESDNet-L} &
{\scriptsize \textbf{Ours}}
\end{tabular}
\caption{Qualitative comparison on UHDM paired benchmark. Our method suppresses moir\'e while retaining visible texture. The lower-row crops show wood grain and fur around the toy's face.}
\label{fig:qual_paired}
\end{figure}
\begin{figure}[!htb]
\centering
\setlength{\tabcolsep}{1pt}
\renewcommand{\arraystretch}{0.6}
\newcolumntype{C}{>{\centering\arraybackslash}m{0.154\textwidth}}
\begin{tabular}{@{}m{1.5em}@{\hspace{2pt}}CCCCCC}
\rotatebox{90}{\scriptsize\shortstack{vivo X200s\\Elevator Display}} &
\includegraphics[width=0.154\textwidth]{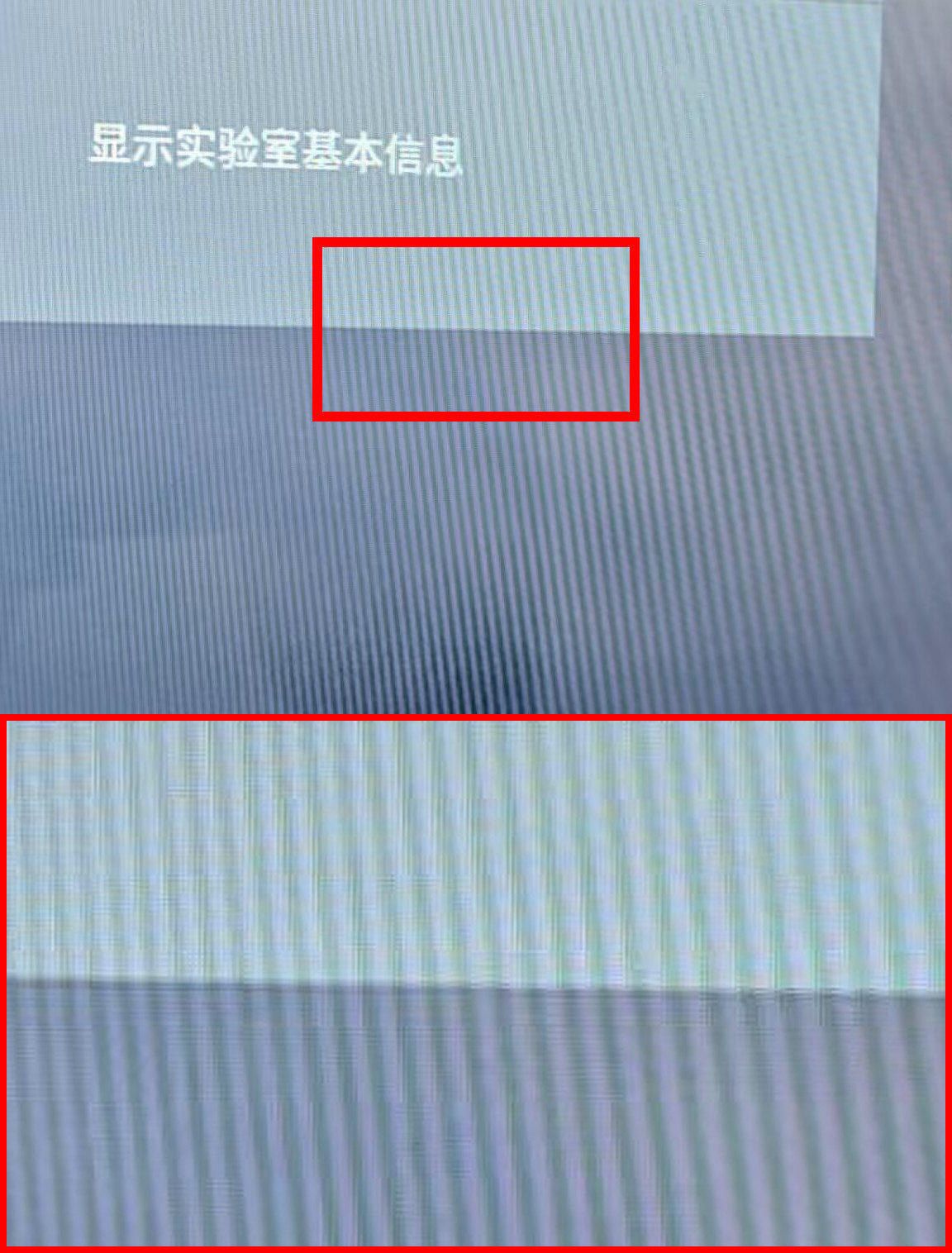} &
\includegraphics[width=0.154\textwidth]{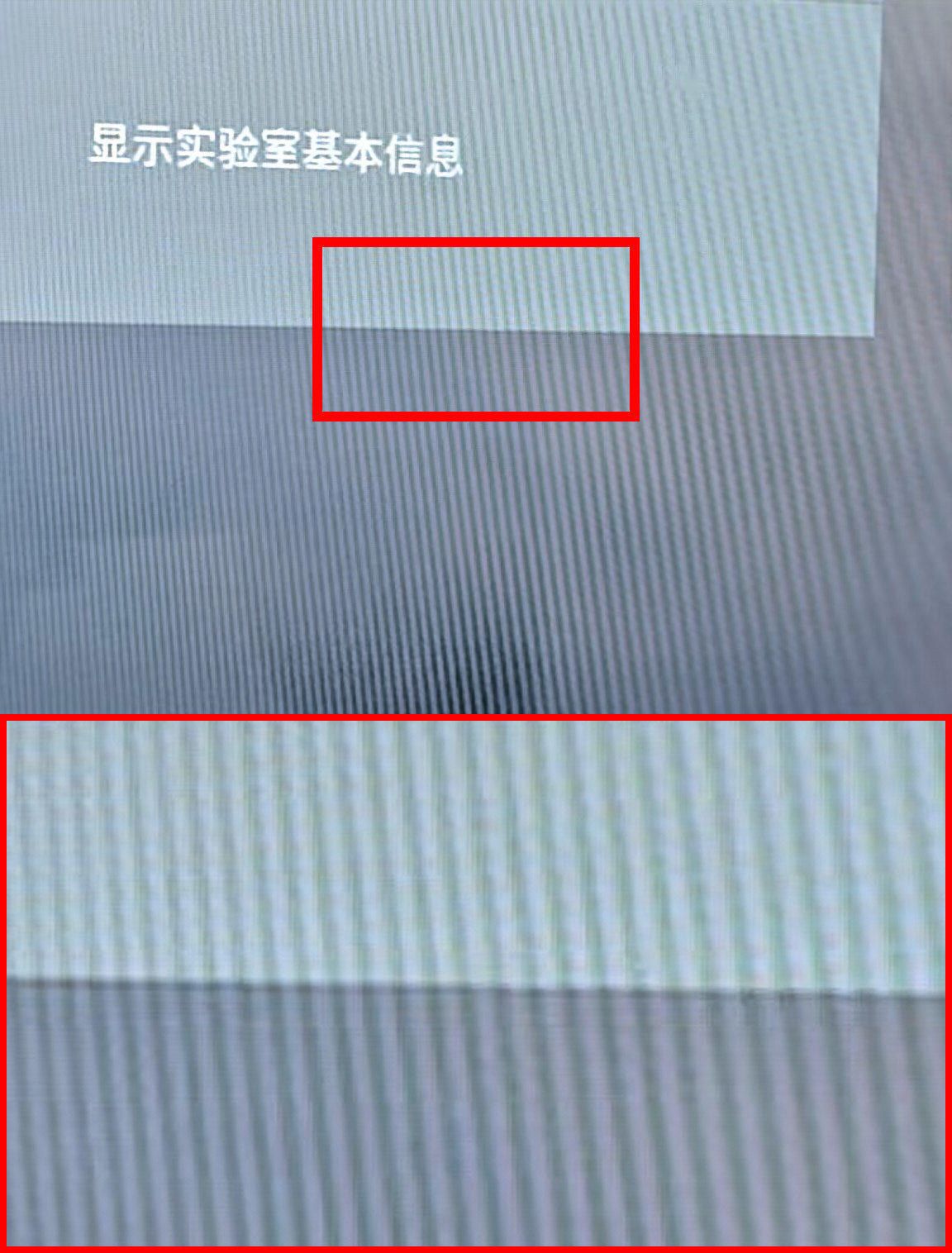} &
\includegraphics[width=0.154\textwidth]{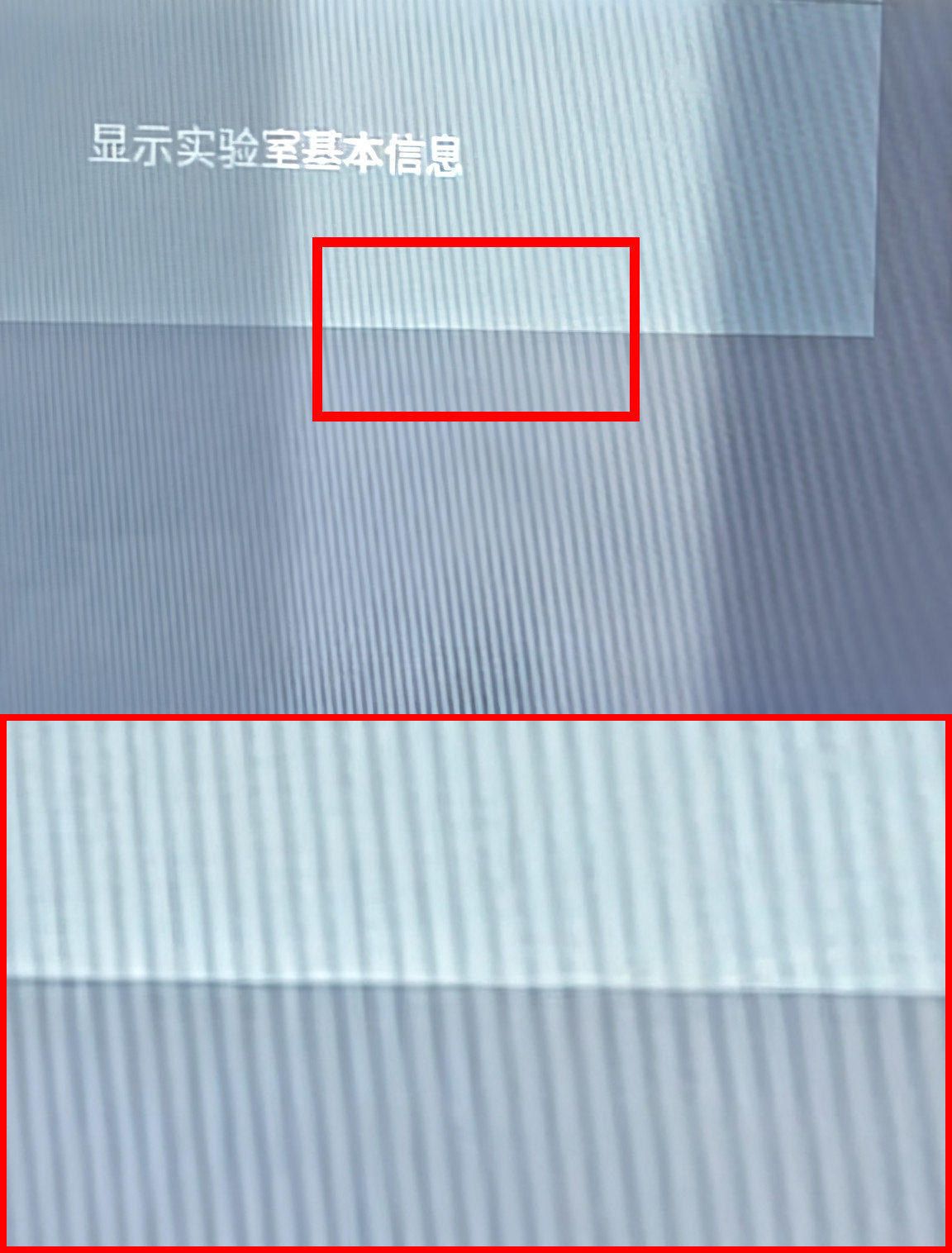} &
\includegraphics[width=0.154\textwidth]{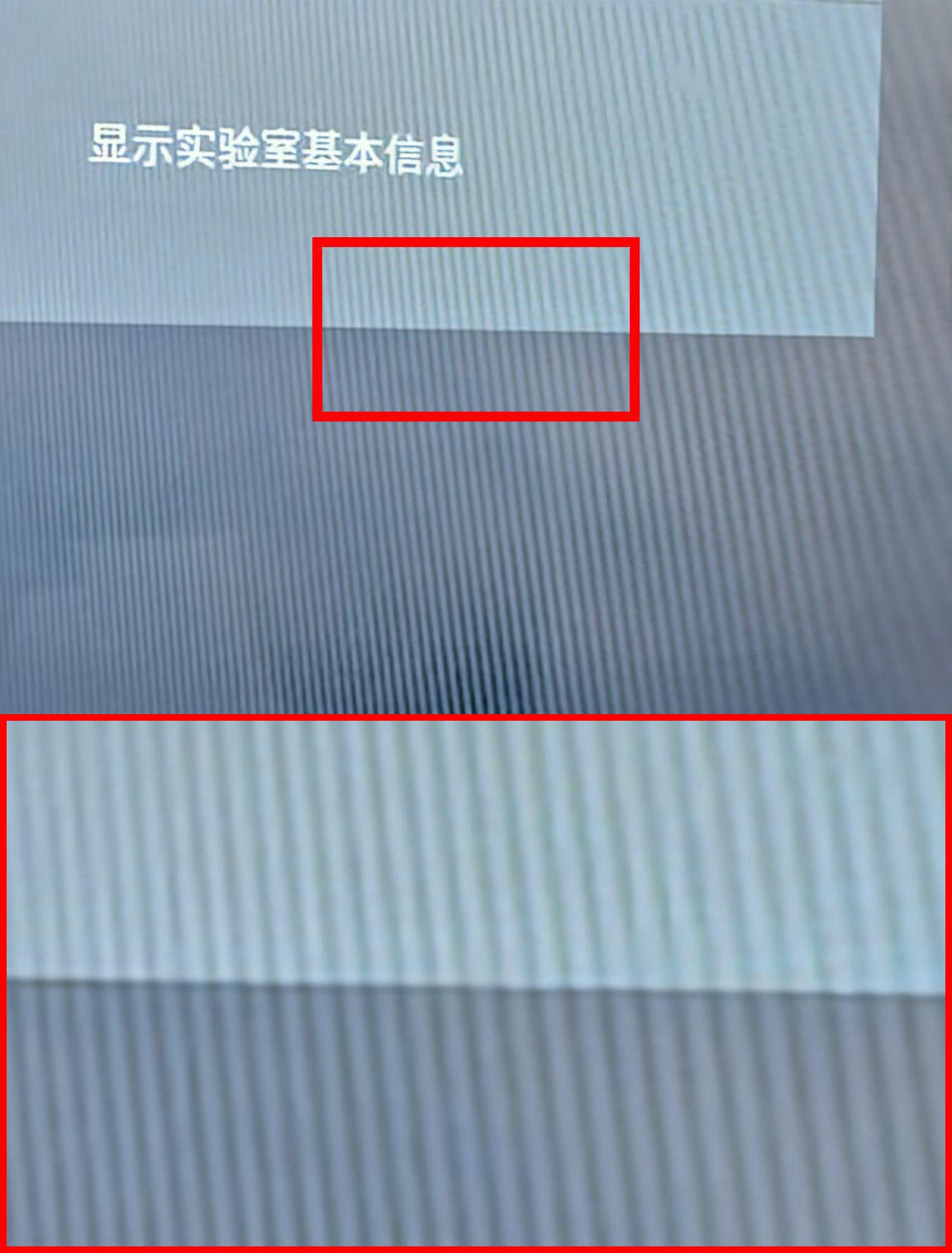} &
\includegraphics[width=0.154\textwidth]{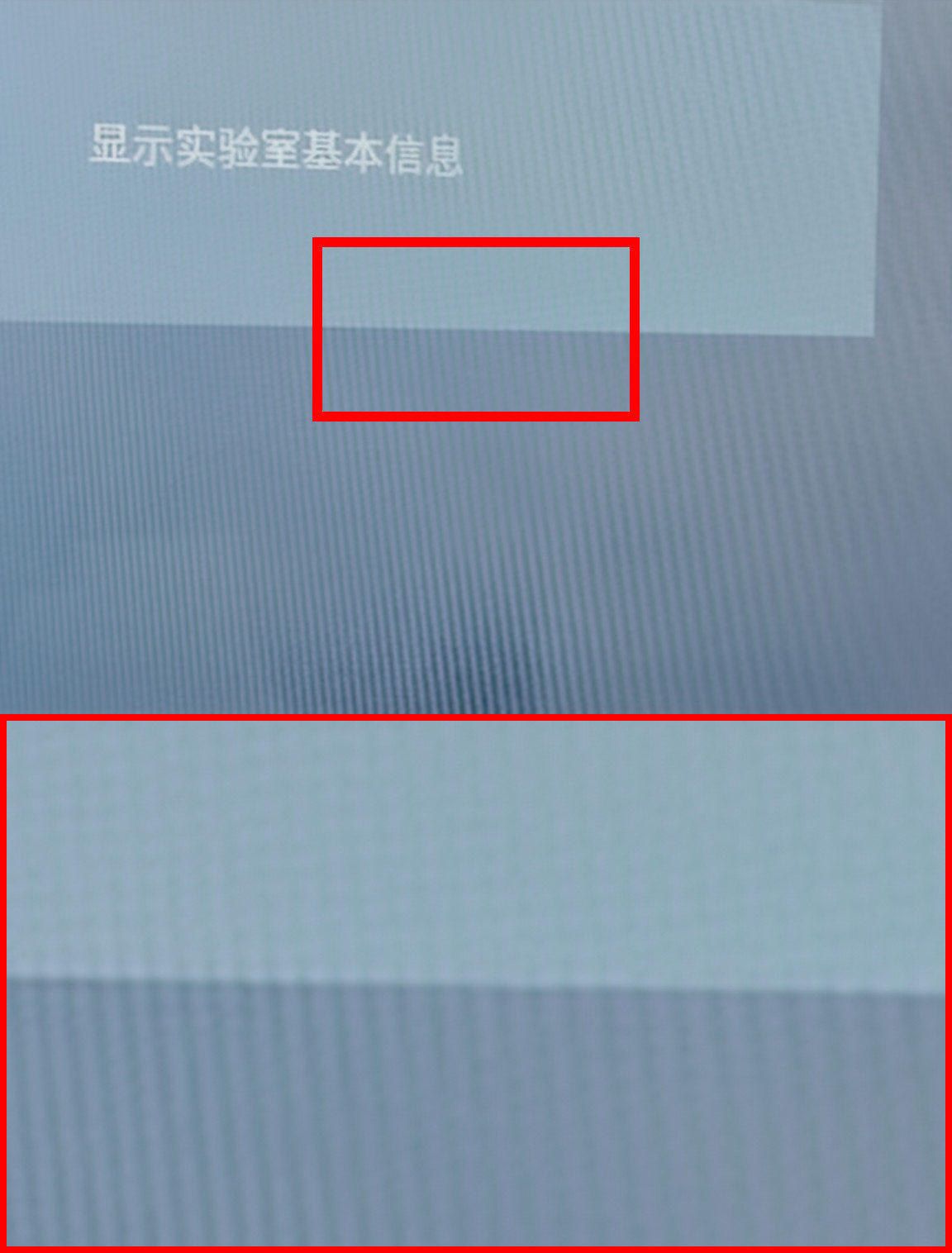} &
\includegraphics[width=0.154\textwidth]{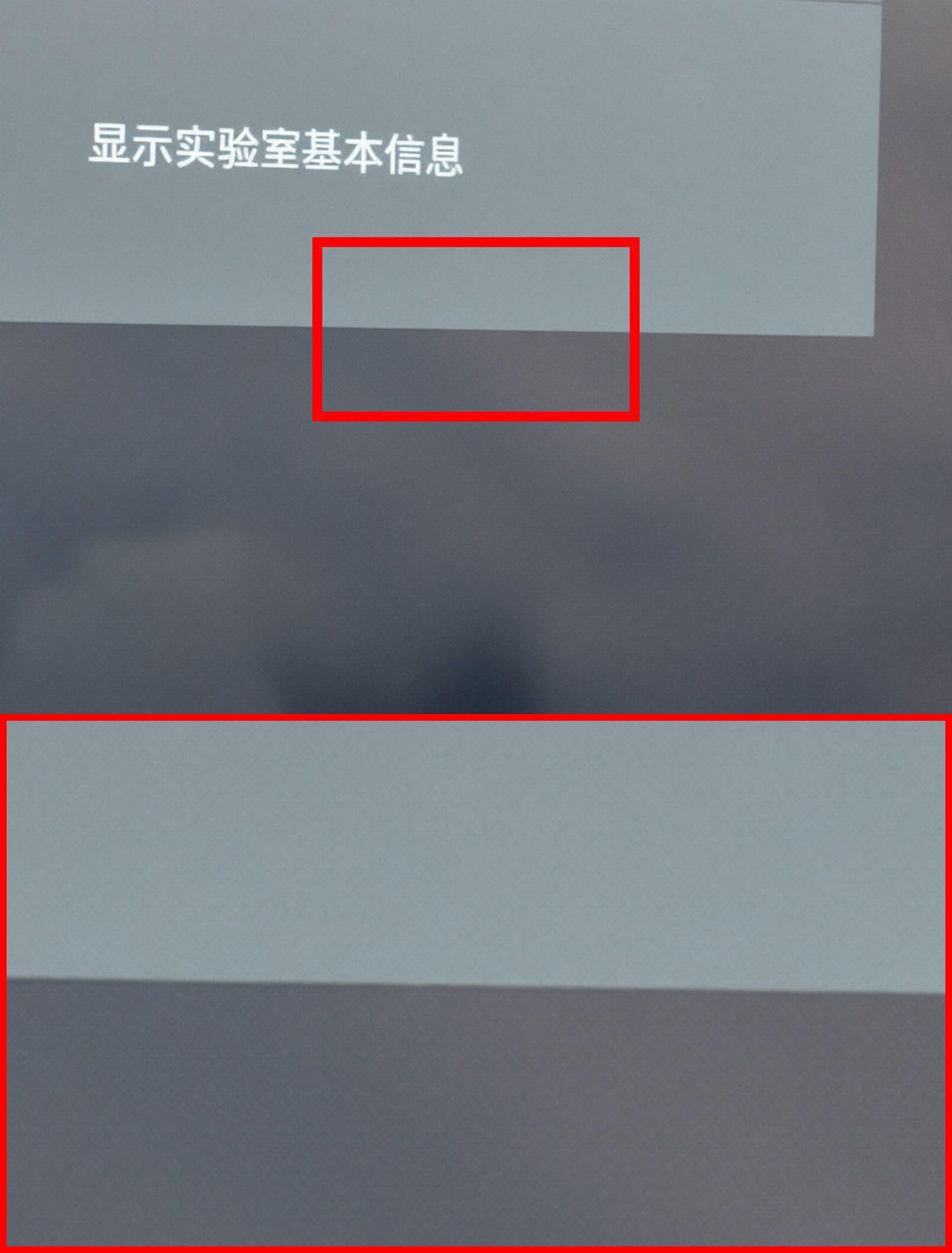} \\
\rotatebox{90}{\scriptsize\shortstack{vivo X200s\\Mall Screen}} &
\includegraphics[width=0.154\textwidth]{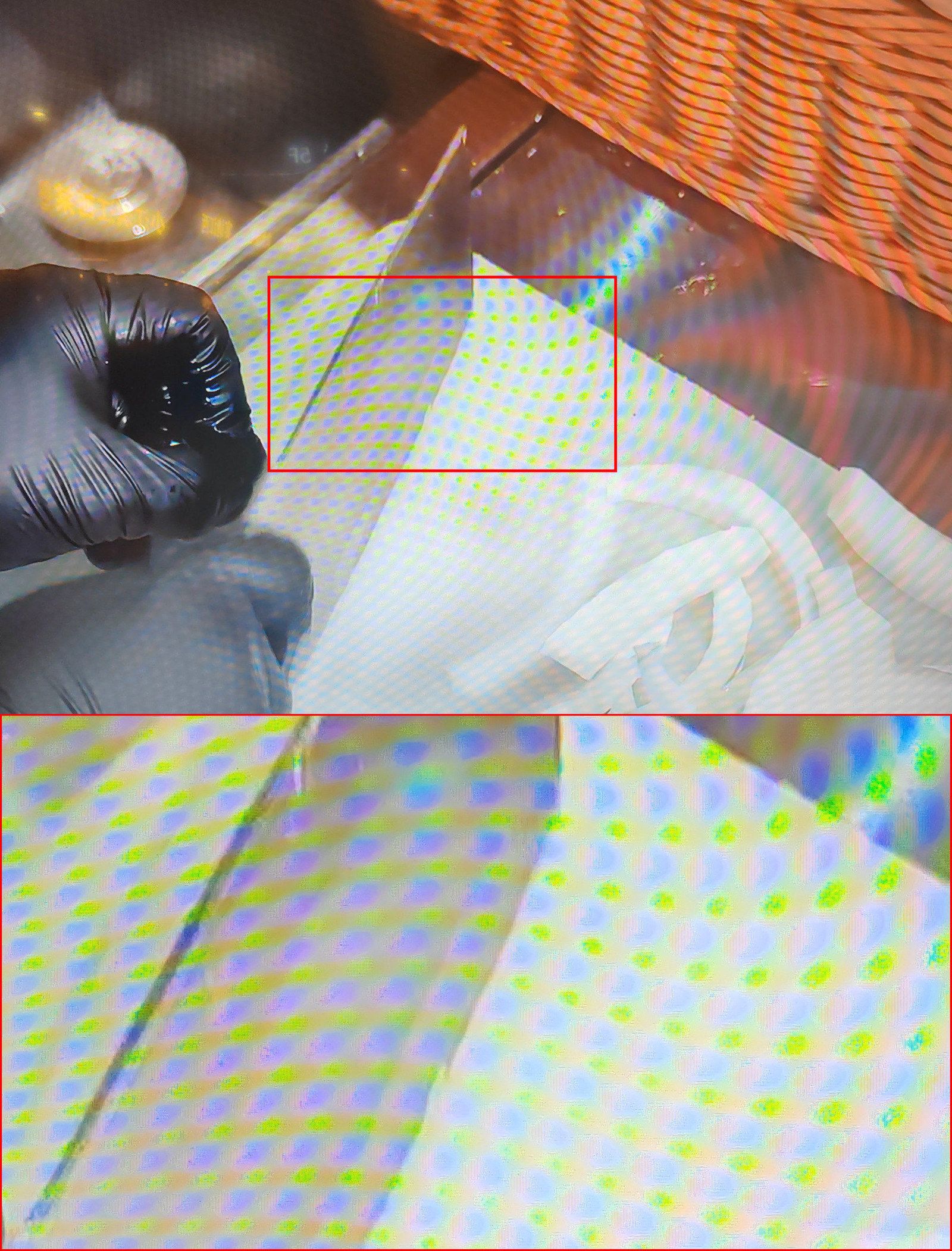} &
\includegraphics[width=0.154\textwidth]{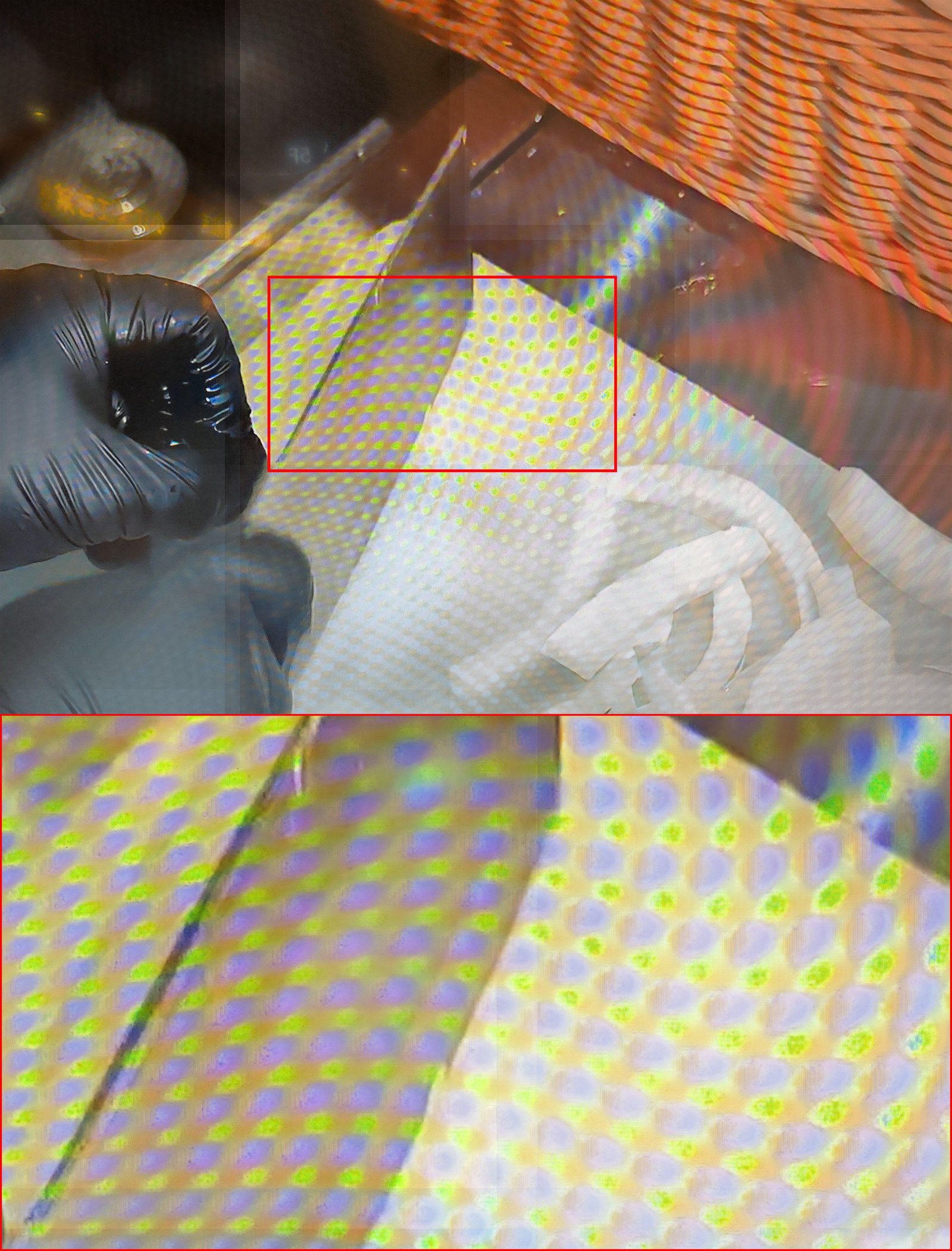} &
\includegraphics[width=0.154\textwidth]{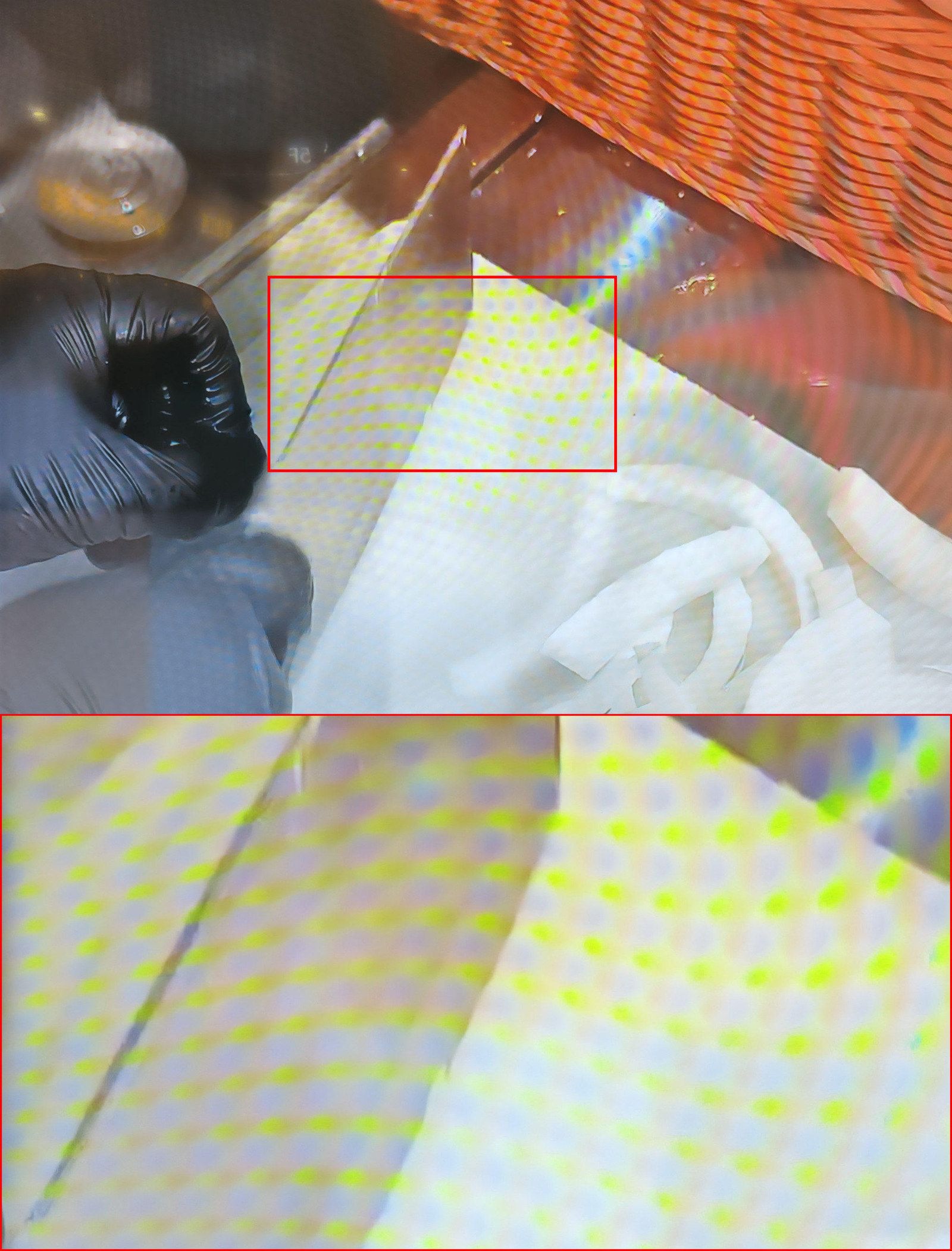} &
\includegraphics[width=0.154\textwidth]{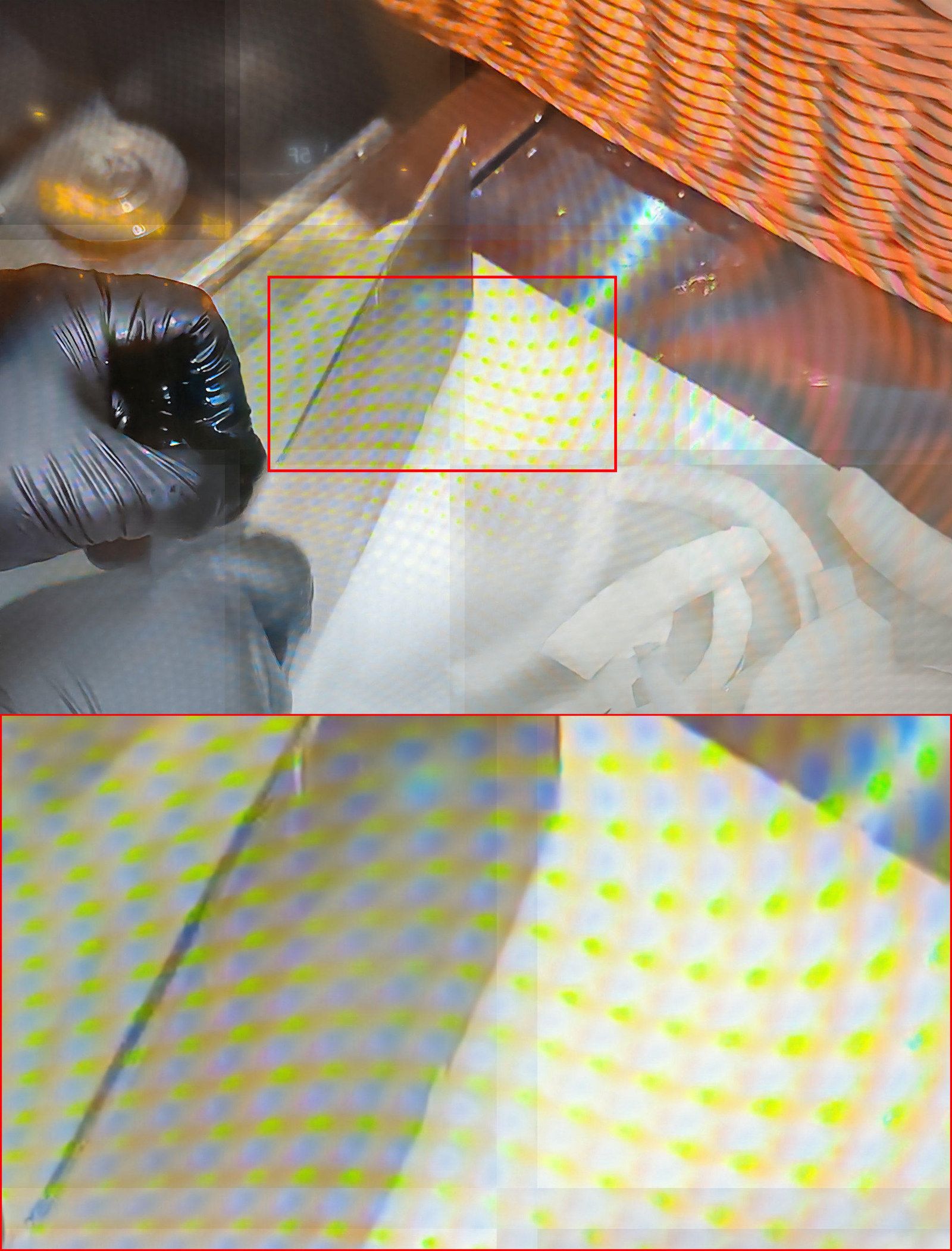} &
\includegraphics[width=0.154\textwidth]{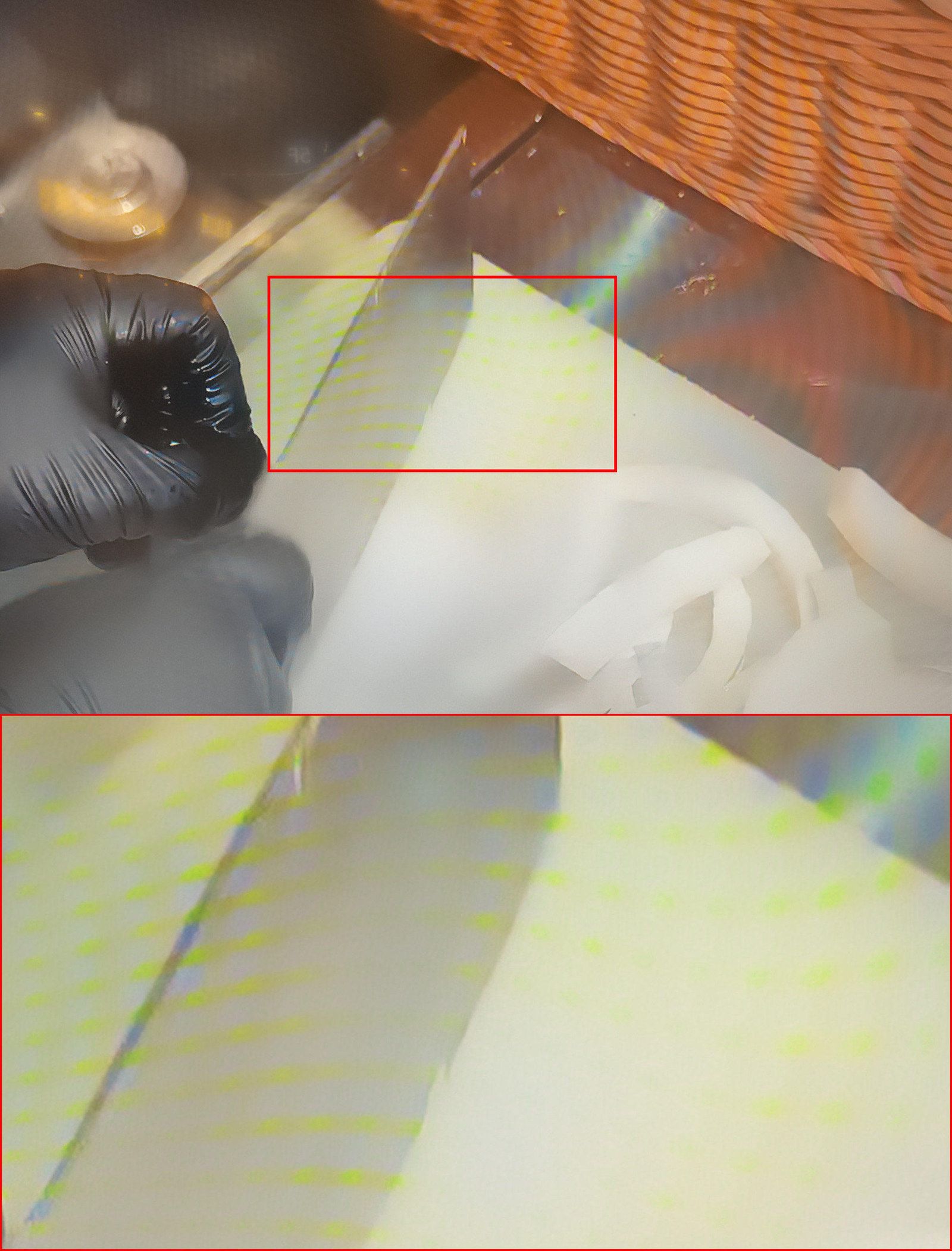} &
\includegraphics[width=0.154\textwidth]{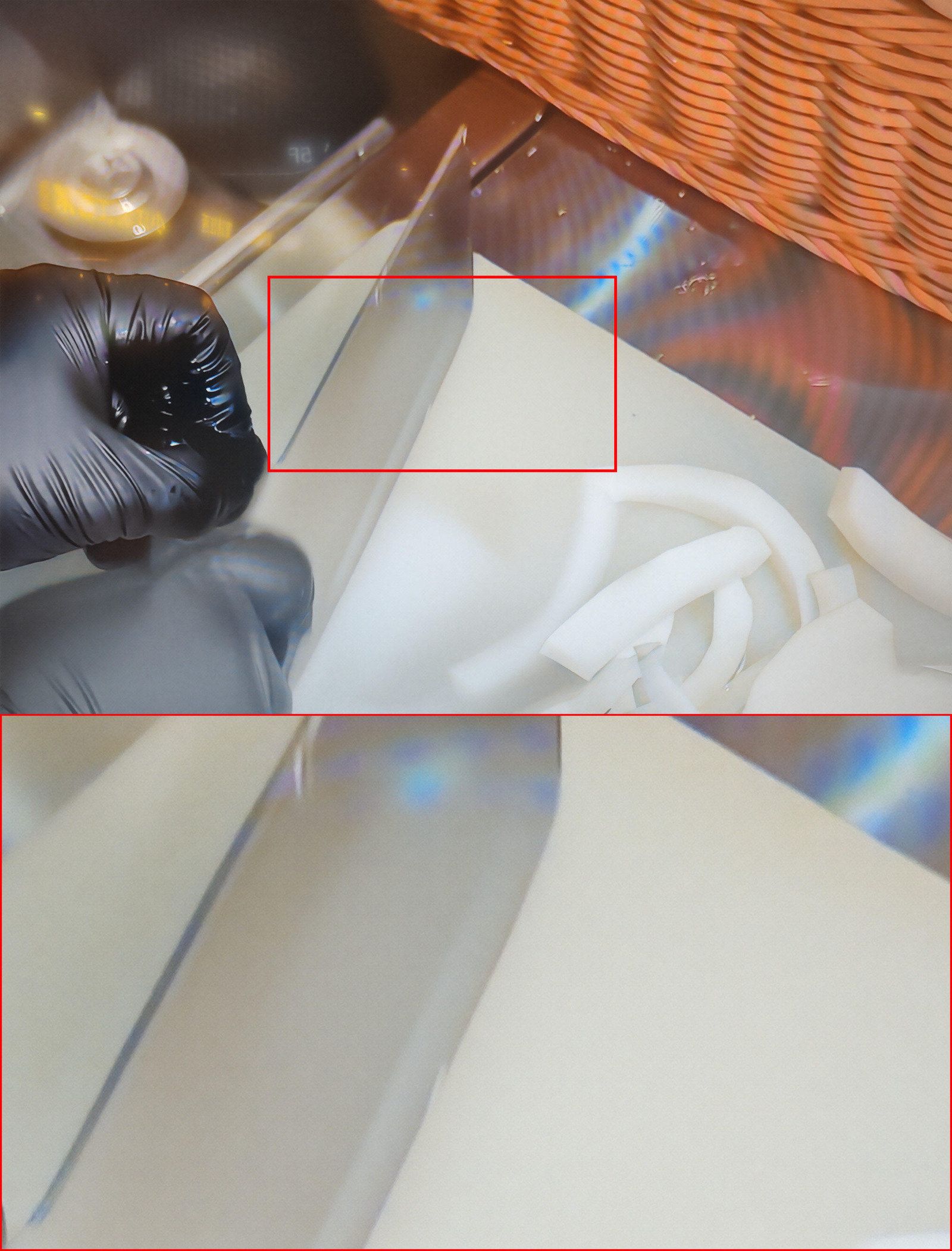} \\
\rotatebox{90}{\scriptsize\shortstack{iPhone 17\\Subway Display}} &
\includegraphics[width=0.154\textwidth]{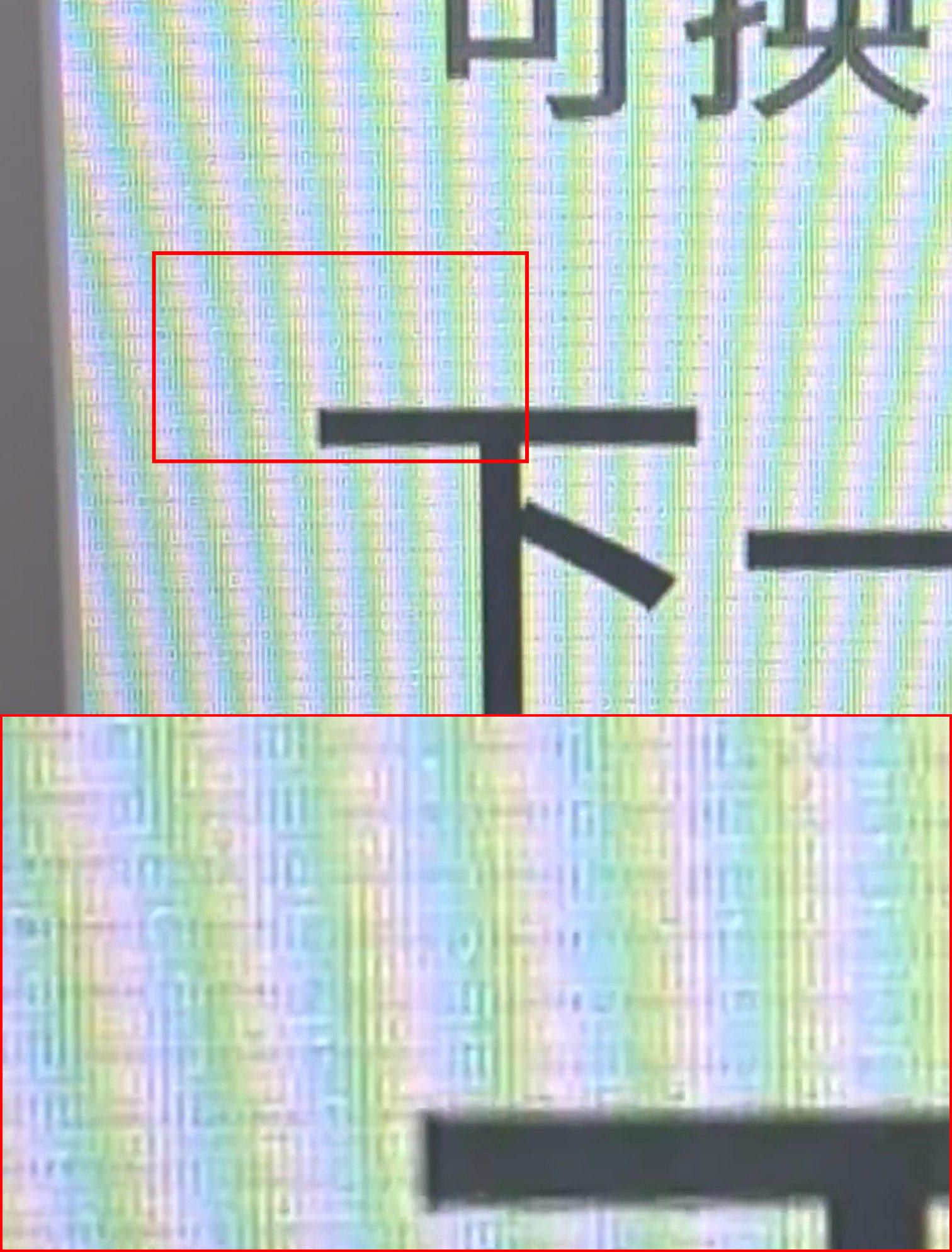} &
\includegraphics[width=0.154\textwidth]{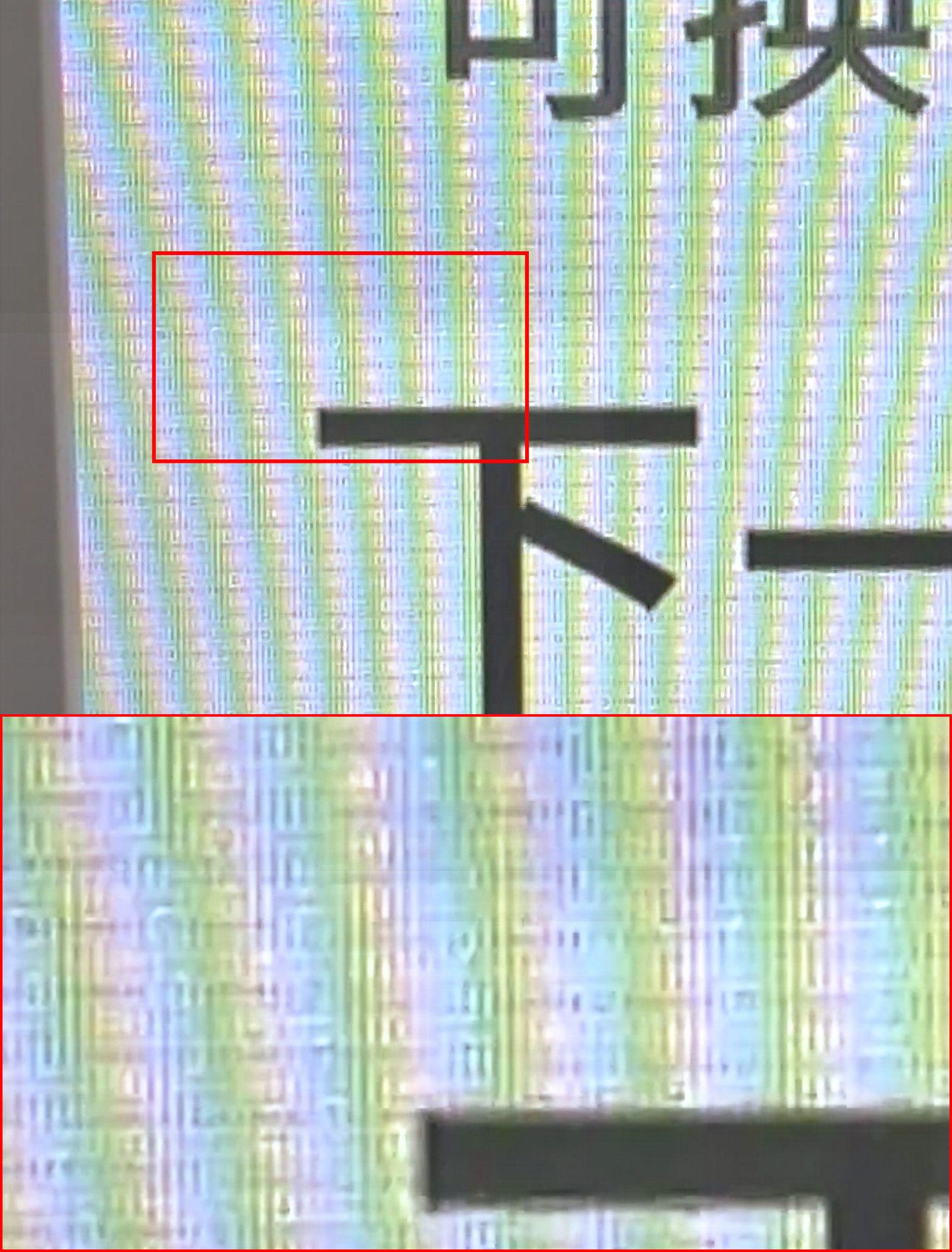} &
\includegraphics[width=0.154\textwidth]{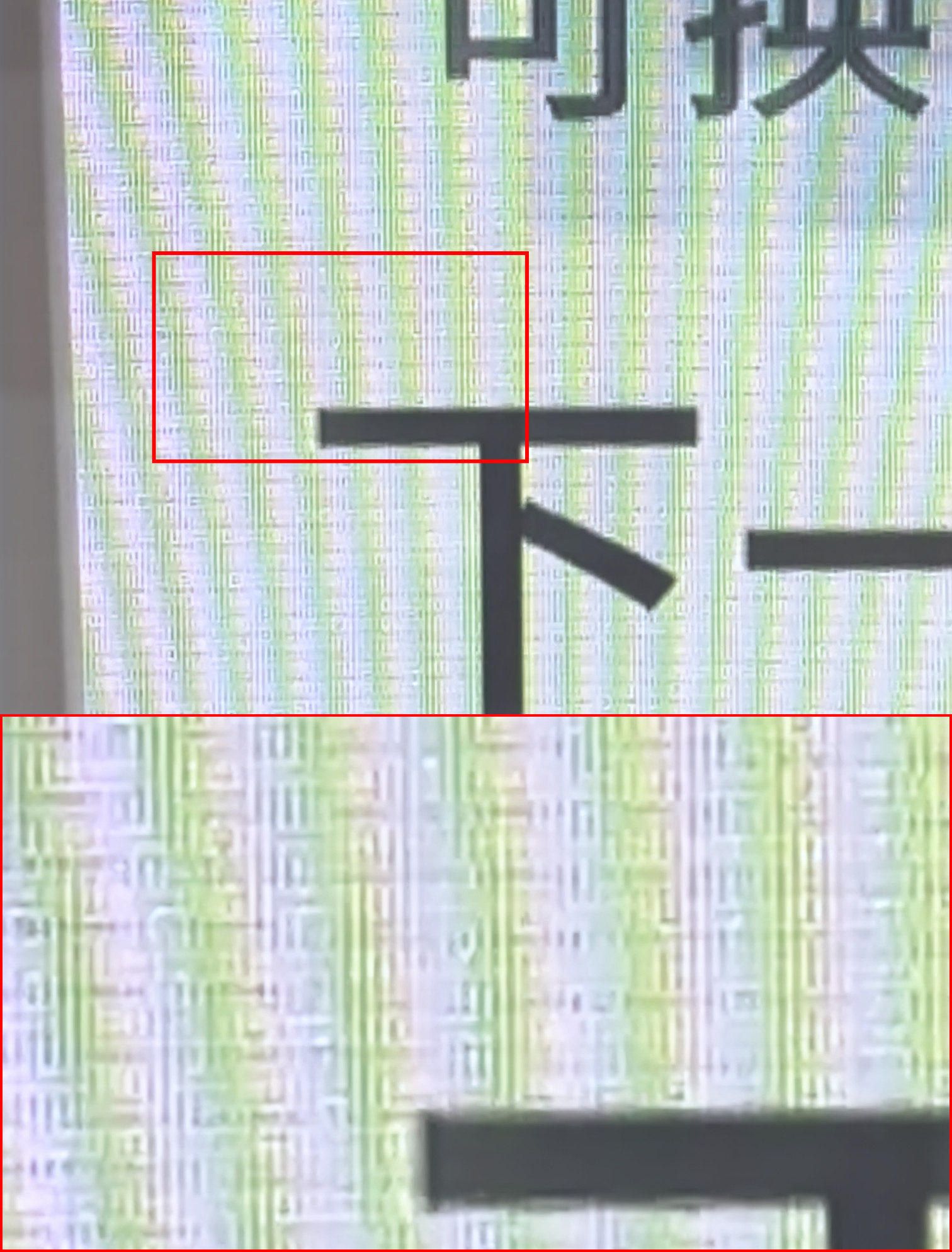} &
\includegraphics[width=0.154\textwidth]{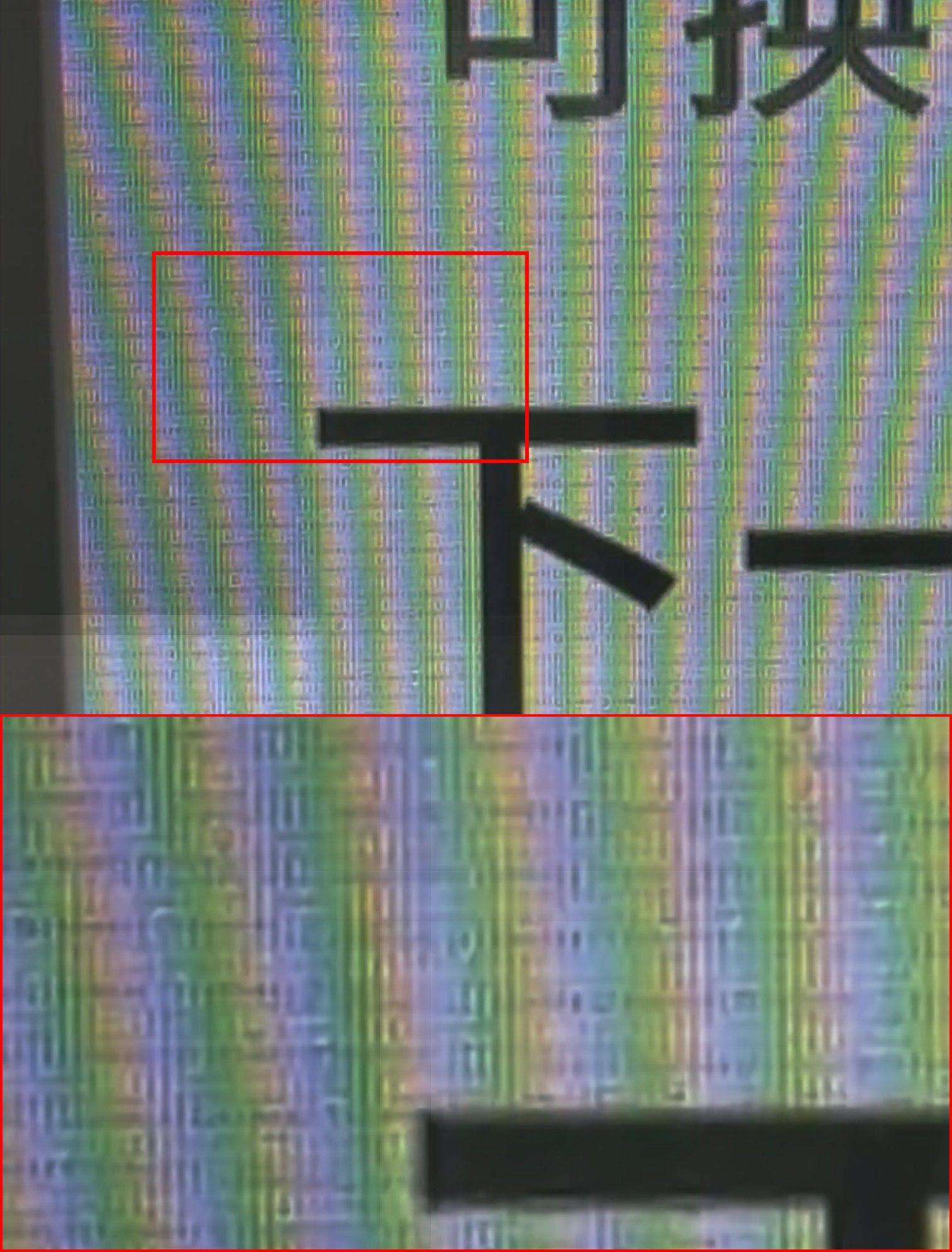} &
\includegraphics[width=0.154\textwidth]{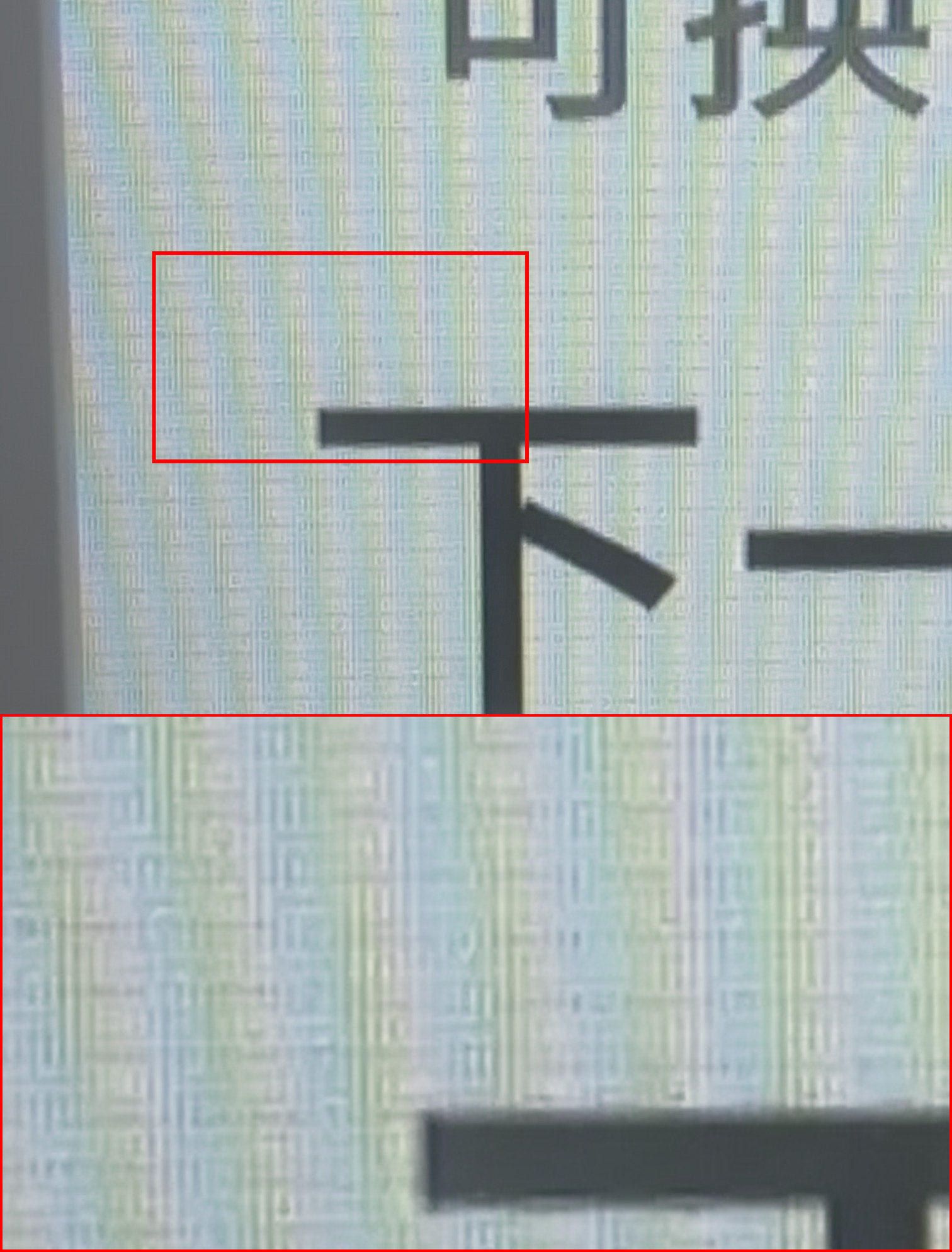} &
\includegraphics[width=0.154\textwidth]{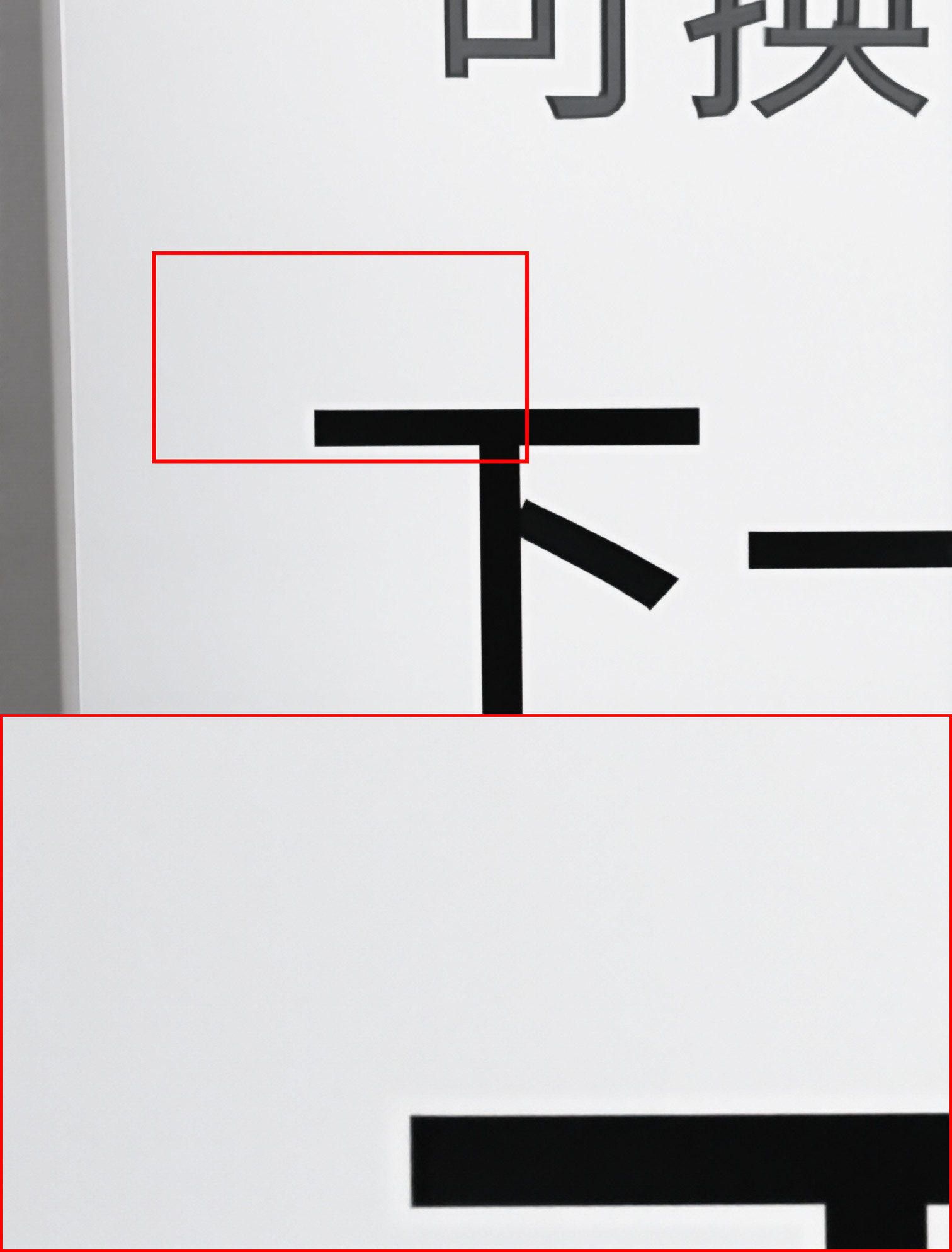} \\
\rotatebox{90}{\scriptsize\shortstack{iPhone 16 Pro Max\\Laptop Display}} &
\includegraphics[width=0.154\textwidth]{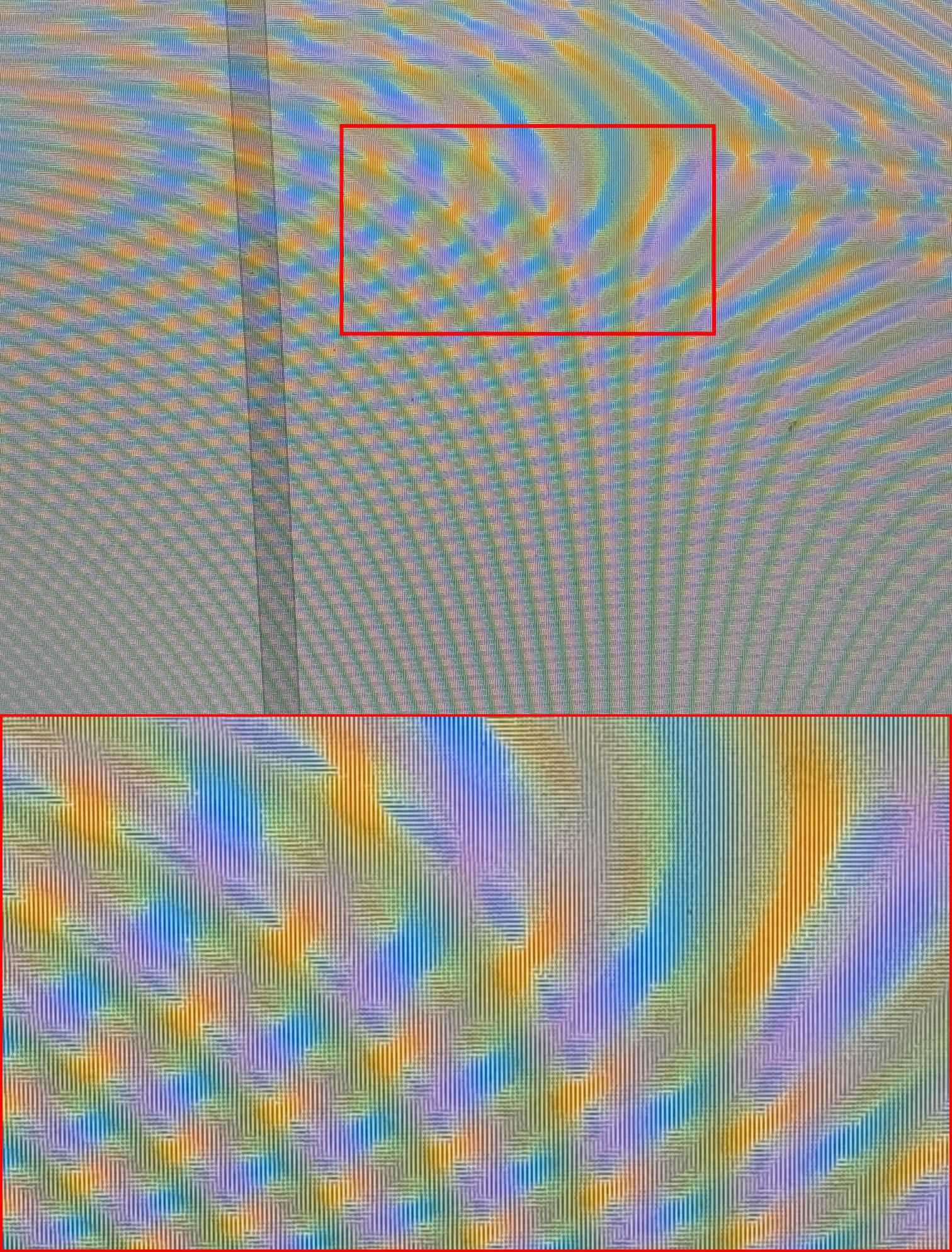} &
\includegraphics[width=0.154\textwidth]{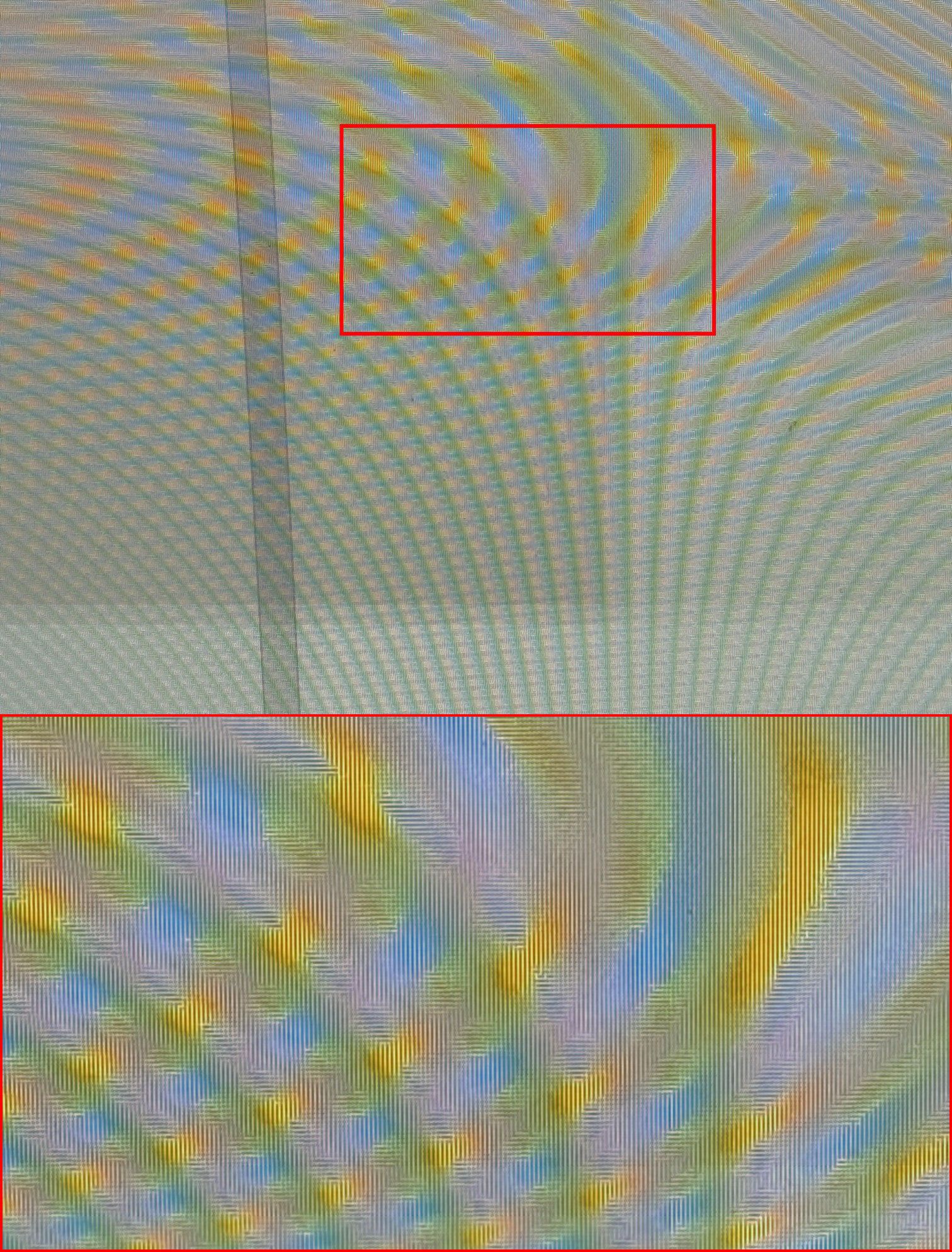} &
\includegraphics[width=0.154\textwidth]{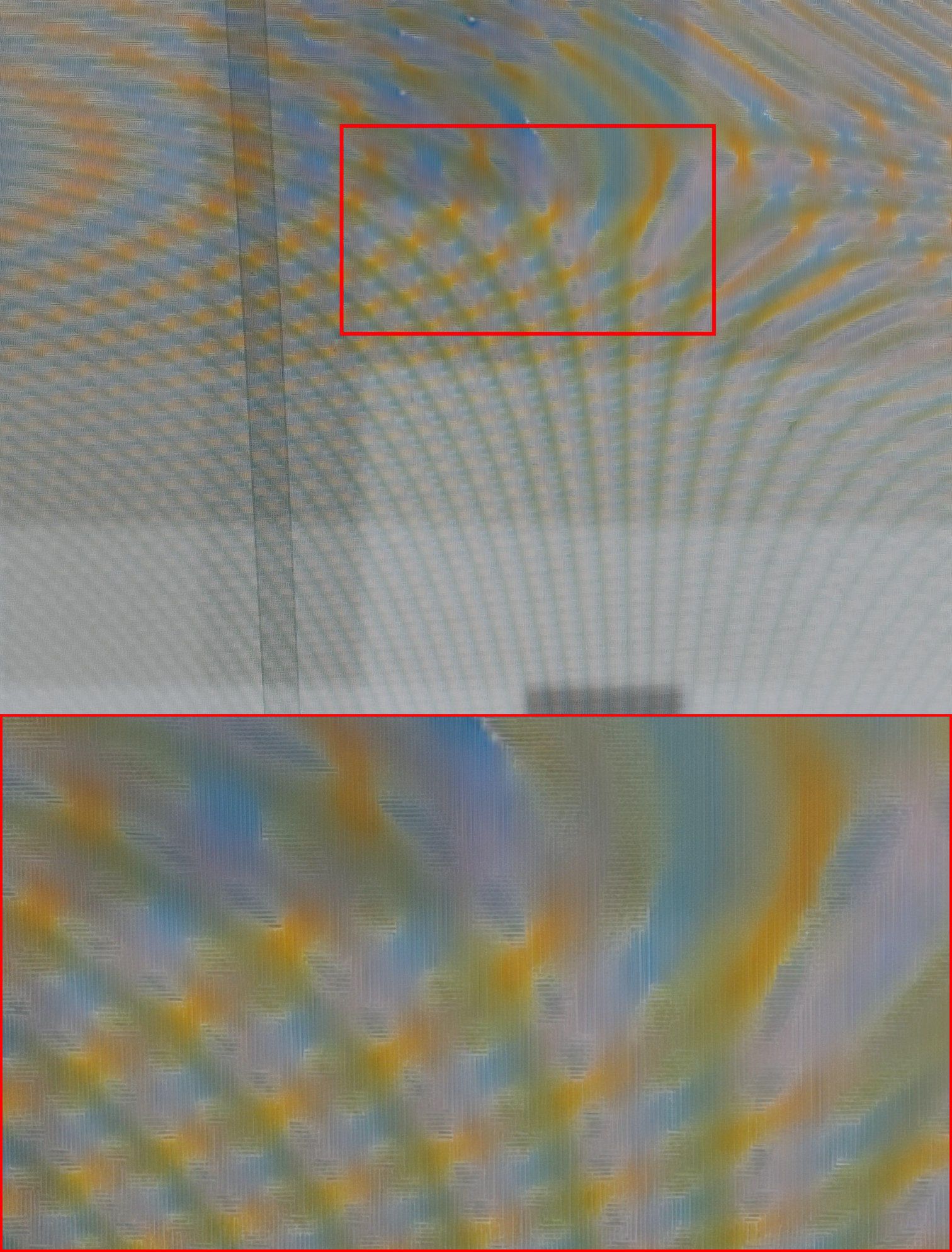} &
\includegraphics[width=0.154\textwidth]{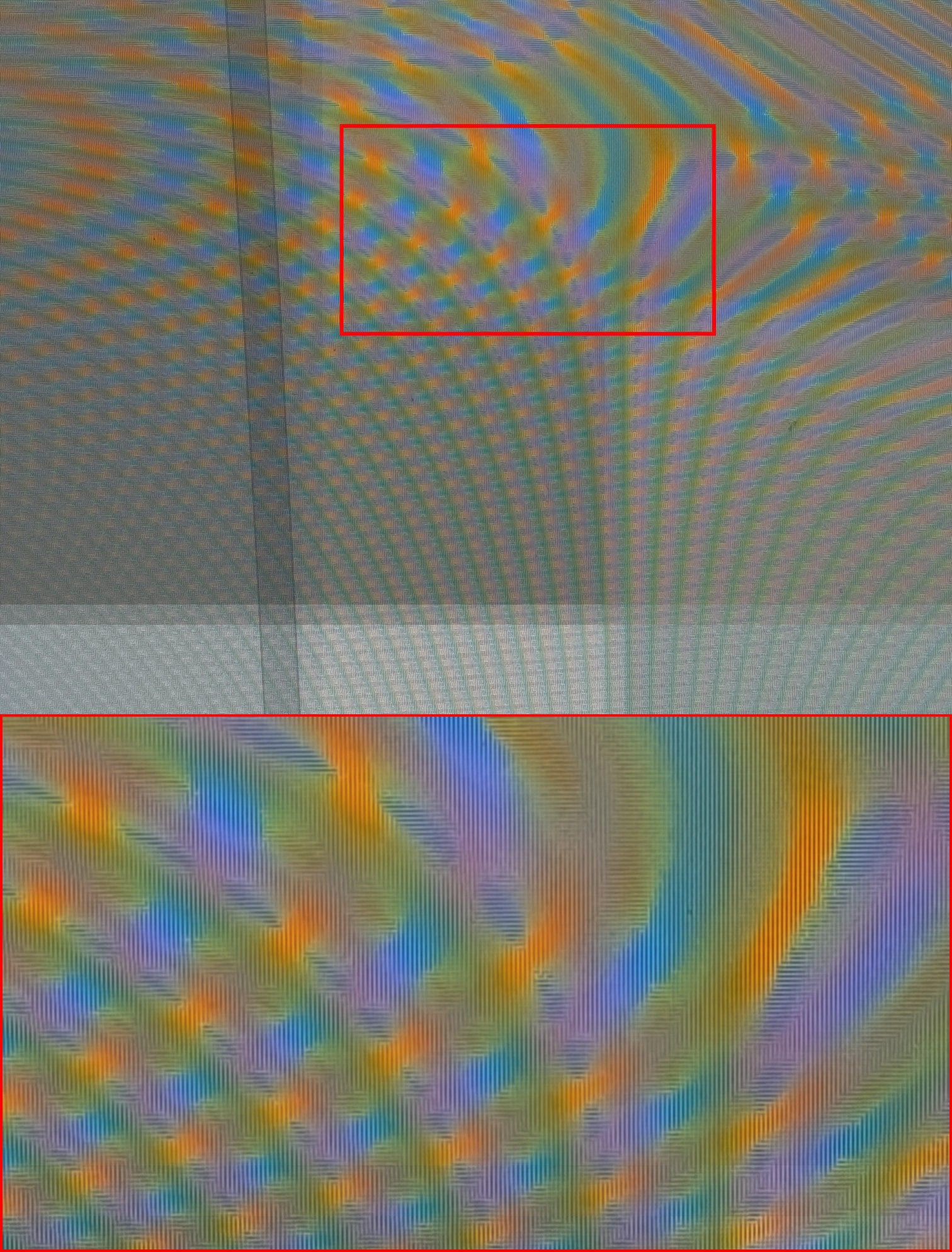} &
\includegraphics[width=0.154\textwidth]{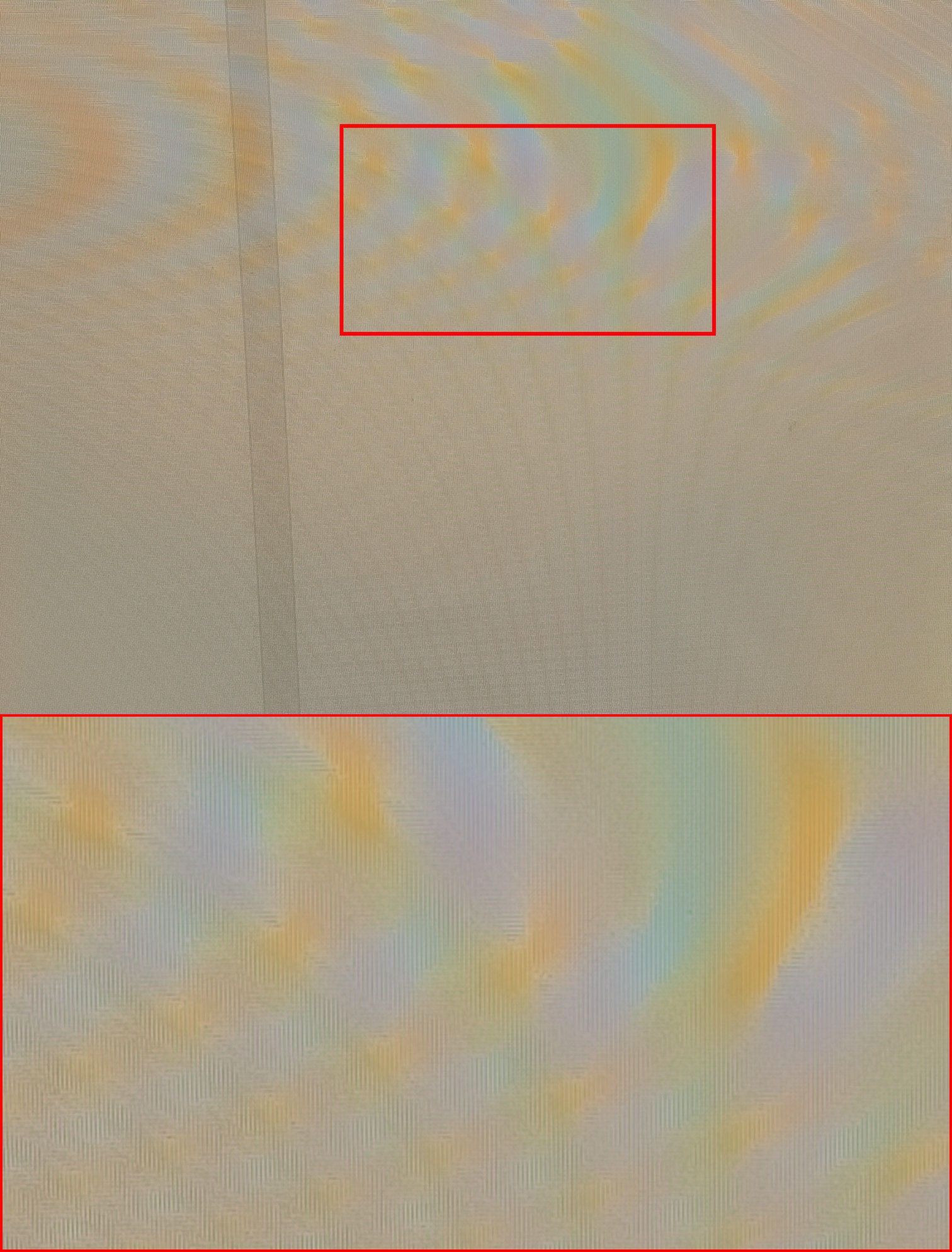} &
\includegraphics[width=0.154\textwidth]{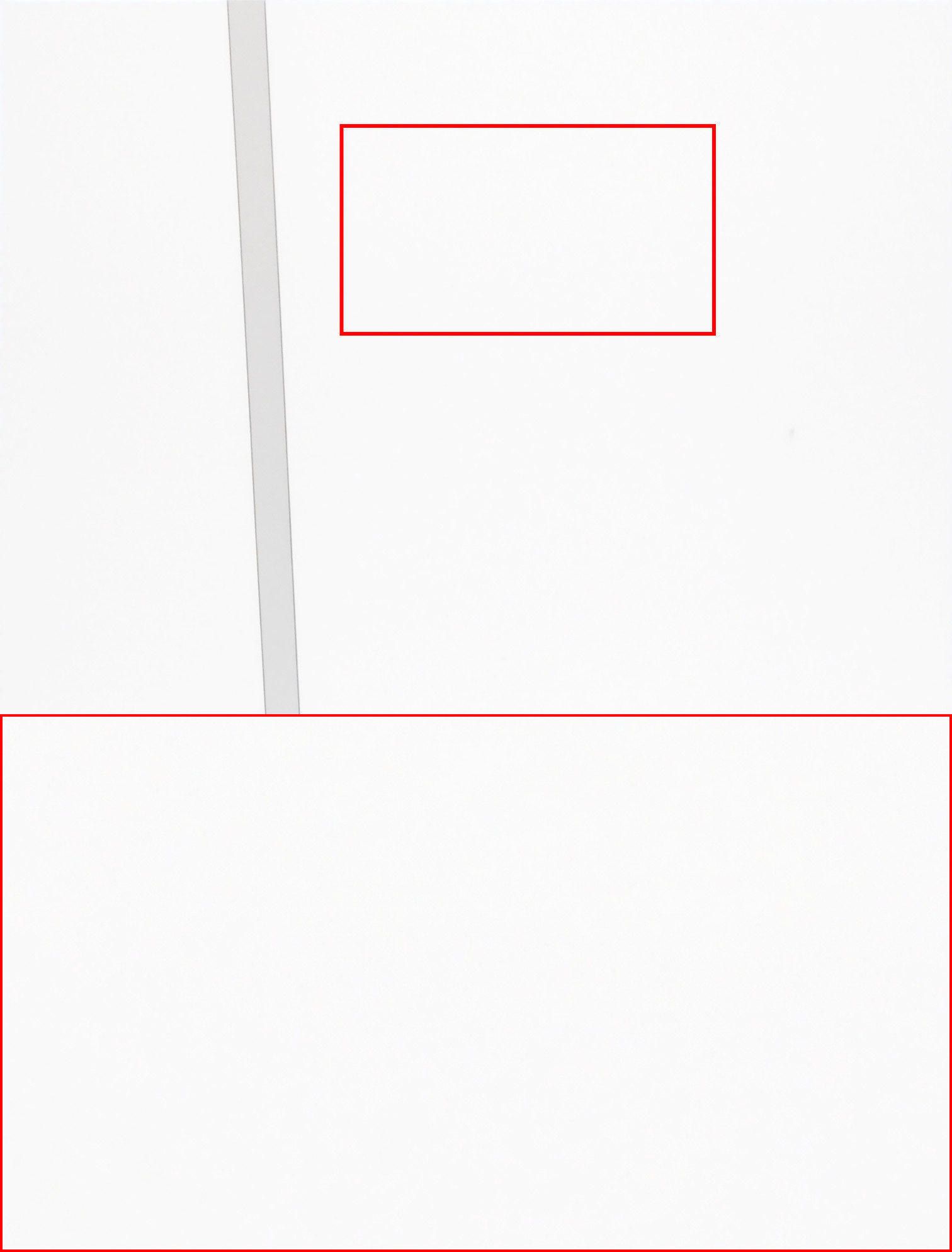} \\[2pt]
&
{\scriptsize Input} &
{\scriptsize AdaIR} &
{\scriptsize OSEDiff} &
{\scriptsize MoCE-IR} &
{\scriptsize ESDNet-L} &
{\scriptsize \textbf{Ours}}
\end{tabular}
\caption{Qualitative results on DailyMoir\'e130. Samples are cropped from the original 4K images for better visualization. These everyday screen captures illustrate challenging mixed-frequency moir\'e.}
\label{fig:qual_real}
\end{figure}
Fig.~\ref{fig:qual_paired} shows qualitative comparisons on paired benchmarks. Several baselines leave visible residual moir\'e on the toy's face. In contrast, our method suppresses these artifacts while retaining visible wood grain and fur around the face. Fig.~\ref{fig:qual_real} further demonstrates generalization to challenging everyday captures in DailyMoir\'e130. The from-scratch baselines shown leave visible artifacts in these out-of-distribution real-world cases. In contrast, our method uses the pretrained editor prior to produce cleaner results, particularly where high-frequency stripes and low-frequency color banding occur within the same image. The DailyMoir\'e130 examples also show faithful text reproduction.

\subsection{Ablation Study}\label{sec:ablations}
\par
\begin{figure}[!htb]
\centering
\setlength{\tabcolsep}{1pt}
\renewcommand{\arraystretch}{0.6}
\begin{tabular}{ccccc}
\includegraphics[width=0.19\textwidth]{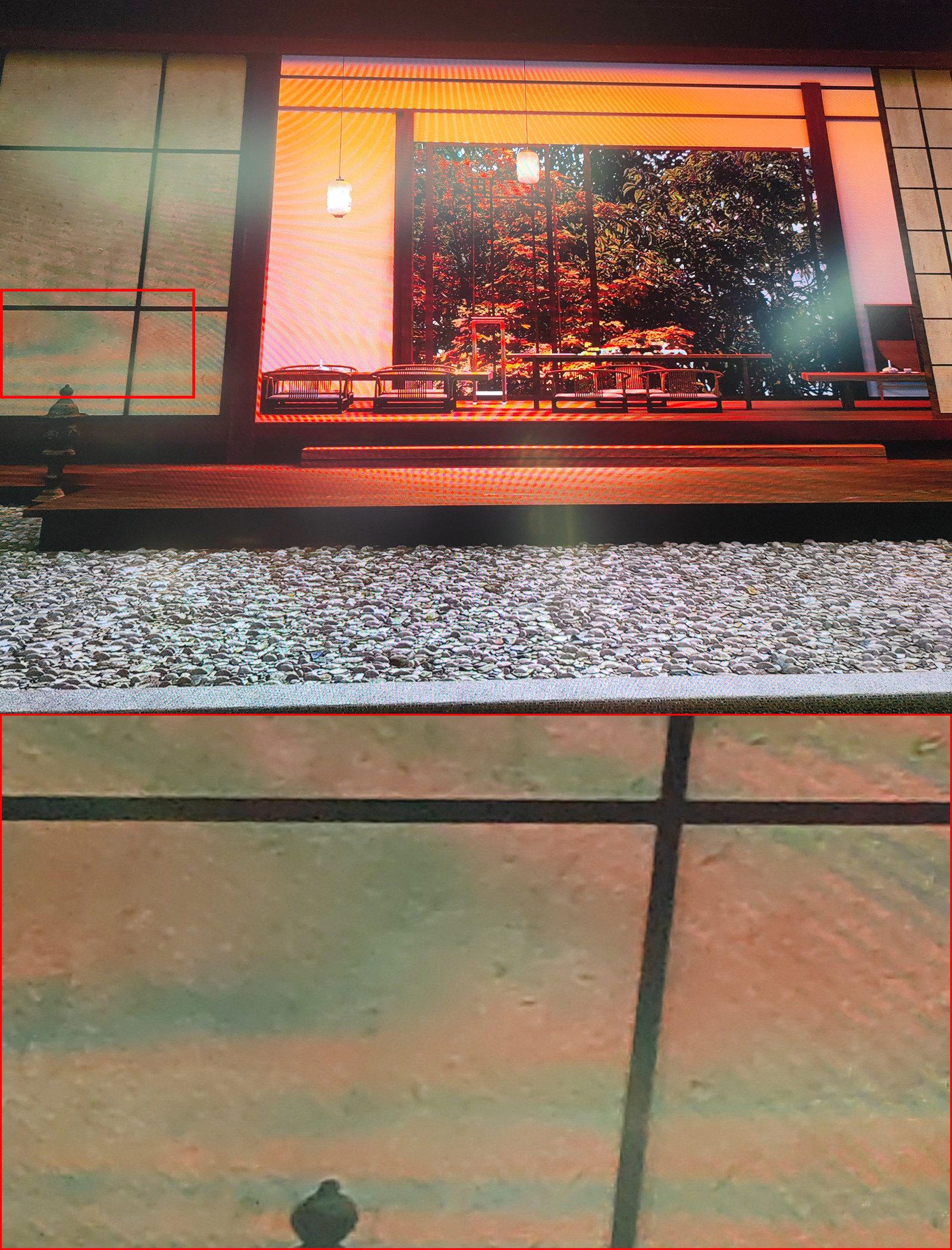} &
\includegraphics[width=0.19\textwidth]{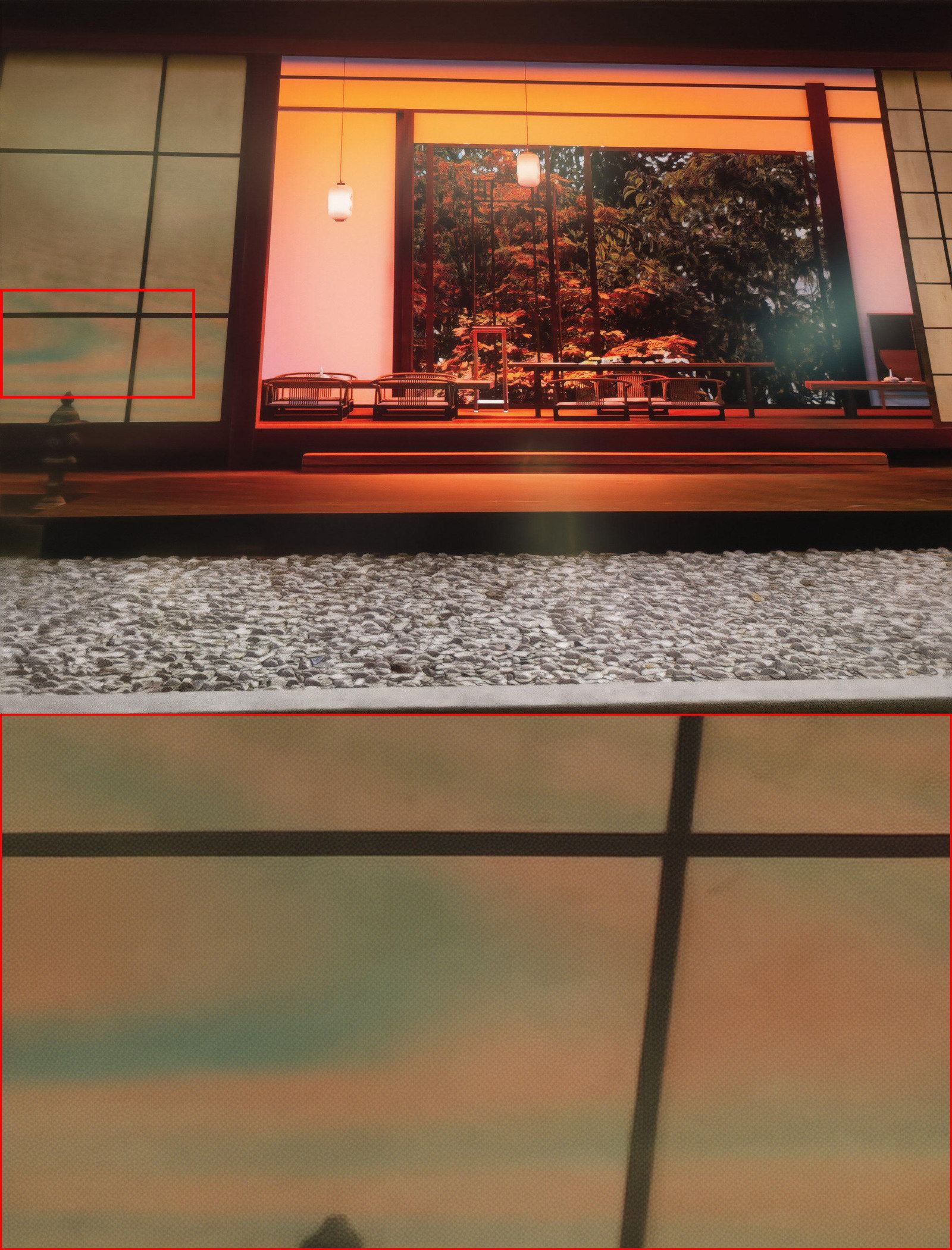} &
\includegraphics[width=0.19\textwidth]{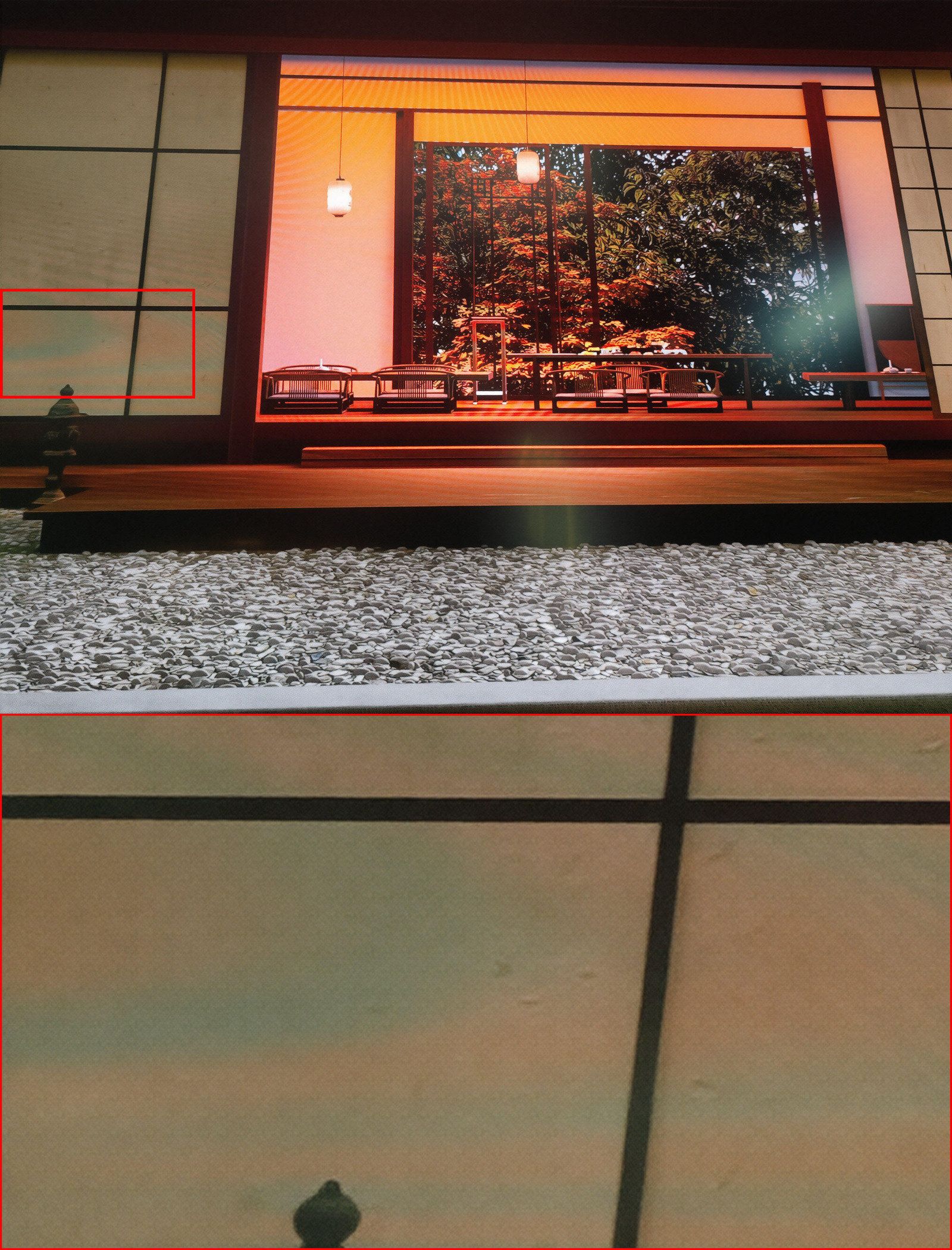} &
\includegraphics[width=0.19\textwidth]{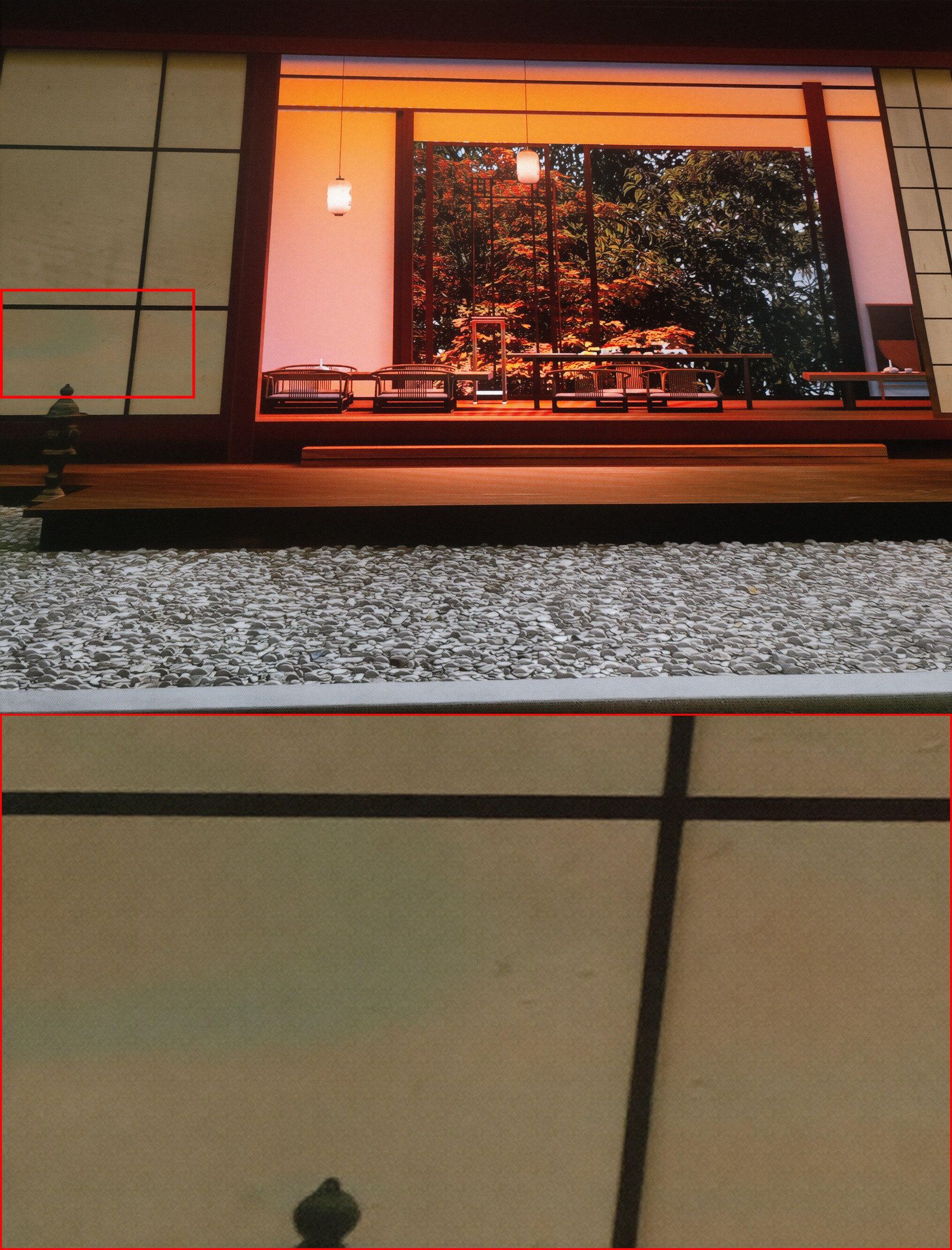} &
\includegraphics[width=0.19\textwidth]{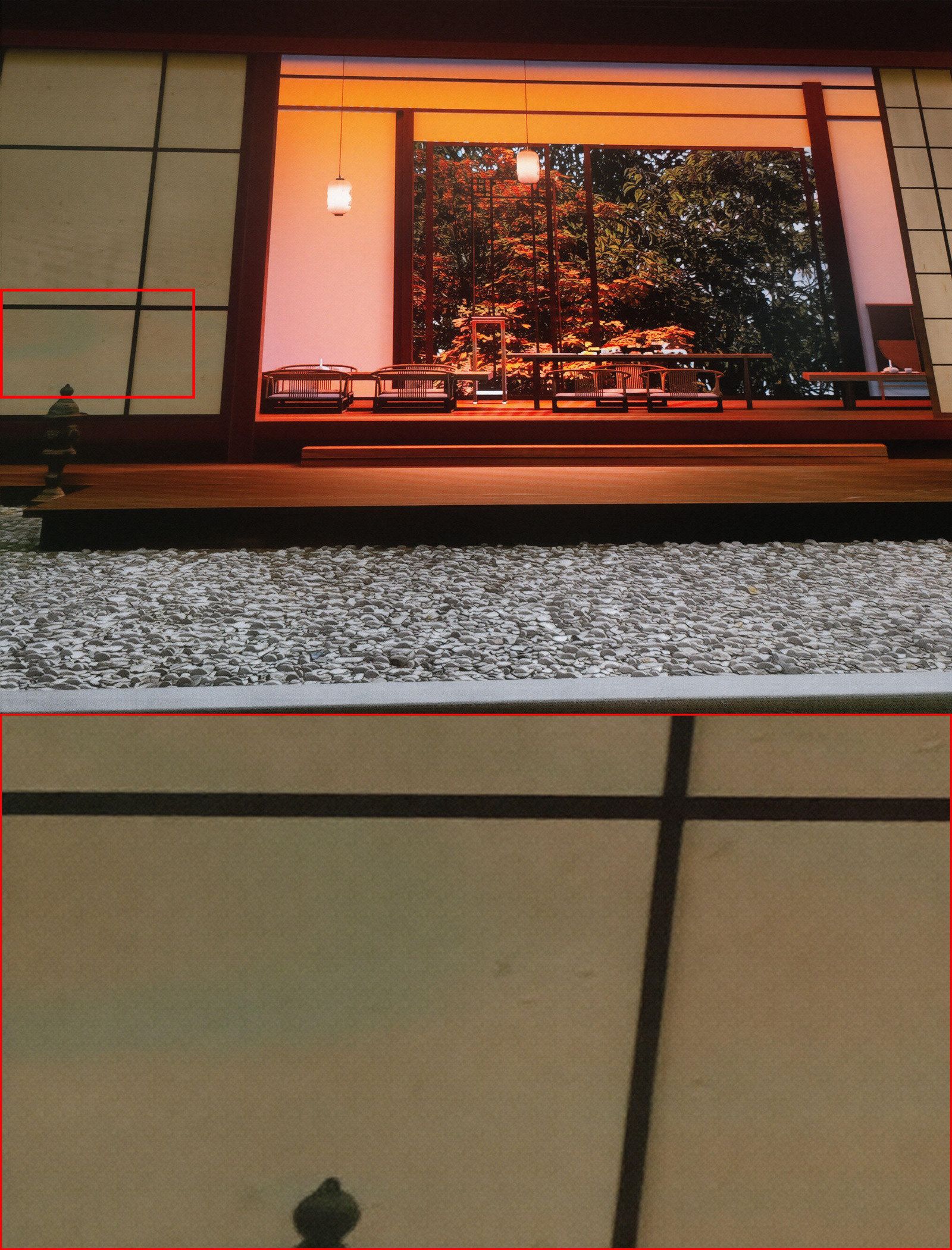} \\[2pt]
{\scriptsize Input} &
{\scriptsize Vanilla FM} &
{\scriptsize w/o condition shift} &
{\scriptsize w/o black image} &
{\scriptsize \textbf{Ours}}
\end{tabular}
\caption{Visual ablation on DailyMoir\'e130. The full model reduces color banding in this example.}
\label{fig:ablation}
\end{figure}

\noindent\begin{minipage}{\linewidth}
\paragraph{Component ablation.}
Our five-step training with endpoint losses improves all six metrics over single-step Vanilla FM (Tab.~\ref{tab:ablation_component}). This comparison measures their joint effect. Keeping the Stage~1 condition throughout Stage~2 ($r=1$) slightly improves PSNR but degrades perceptual quality and moir\'e suppression. Black image conditioning trades marginal RMC/RMA for better LPIPS, SSIM, and MUSIQ. Fig.~\ref{fig:ablation} illustrates these effects on DailyMoir\'e130.
\end{minipage}\par

\begin{table}[!htb]
\centering
\caption{Training and inference ablations on UHDM.}\label{tab:ablation_component}
\papertablestyle
\begin{tabular*}{\linewidth}{@{\extracolsep{\fill}}lrrrrrr@{}}
\toprule
Setting & PSNR$\uparrow$ & SSIM$\uparrow$ & LPIPS$\downarrow$ & RMC$\downarrow$ & RMA$\downarrow$ & MUSIQ$\uparrow$ \\
\midrule
Vanilla FM (1 step) & 17.51 & 0.669 & 0.195 & 0.0162 & 0.0953 & 65.29 \\
w/o condition shift & \textbf{20.24} & \underline{0.751} & \underline{0.189} & 0.0047 & 0.0255 & \underline{66.84} \\
w/o black image     & \underline{20.00} & 0.737 & \underline{0.189} & \textbf{0.0036} & \textbf{0.0195} & 66.81 \\
\textbf{Ours}   & 19.99 & \textbf{0.789} & \textbf{0.134} & \underline{0.0039} & \underline{0.0212} & \textbf{67.43} \\
\bottomrule
\end{tabular*}
\end{table}

\par\noindent\begin{minipage}{\linewidth}
\paragraph{Condition-shift ratio $r$.}
Tab.~\ref{tab:ablation_r} varies when Stage~2 shifts from the Stage~1 output to the original input. Larger $r$ reduces residual moir\'e but suppresses detail. At $r=0$, Stage~1 is omitted and Stage~2 uses the original input throughout. Compared with this setting, $r=0.3$ reduces RMA by 17.2\% and RMC by 17.0\%, with 0.03 dB lower PSNR. We therefore use $r=0.3$ to balance suppression and detail preservation.
\end{minipage}\par

\begin{table}[!htb]
\centering
\caption{Condition-shift ablation on UHDM, including Stage-2 only ($r=0$).}\label{tab:ablation_r}
\papertablestyle
\begin{tabular*}{\linewidth}{@{\extracolsep{\fill}}lrrrrrr@{}}
\toprule
$r$ & PSNR$\uparrow$ & SSIM$\uparrow$ & LPIPS$\downarrow$ & RMC$\downarrow$ & RMA$\downarrow$ & MUSIQ$\uparrow$ \\\midrule
0.7        & 19.81 & 0.727 & 0.193 & \textbf{0.0032} & \textbf{0.0170} & 66.89 \\
0.5        & 19.88 & 0.732 & 0.191 & \underline{0.0036} & \underline{0.0192} & 66.87 \\
\textbf{0.3 (ours)} & \underline{19.99} & \underline{0.789} & \textbf{0.134} & 0.0039 & 0.0212 & \textbf{67.43} \\
0 (Stage-2 only) & \textbf{20.02} & \textbf{0.793} & \underline{0.135} & 0.0047 & 0.0256 & \underline{67.42} \\

\bottomrule
\end{tabular*}
\end{table}

\par\noindent\begin{minipage}{\linewidth}
\paragraph{Individual endpoint losses.}
We train variants without L1, LPIPS, or FFL while keeping the remaining settings fixed. Removing each loss degrades all six metrics (Tab.~\ref{tab:loss}). Without L1, PSNR falls to 19.51 dB. Removing LPIPS increases LPIPS to 0.145, while removing FFL increases RMA to 0.0308. These results support complementary pixel, perceptual, and frequency supervision.
\par\vspace{6pt}
\captionsetup{hypcap=false}
\centering
\captionof{table}{Individual endpoint-loss ablations on UHDM. All variants retain velocity supervision and the other endpoint losses. Best values are bold.}
\label{tab:loss}
\papertablestyle
\begin{tabular*}{\linewidth}{@{\extracolsep{\fill}}lrrrrrr@{}}
\toprule
Variant & PSNR$\uparrow$ & SSIM$\uparrow$ & LPIPS$\downarrow$ & RMC$\downarrow$ & RMA$\downarrow$ & MUSIQ$\uparrow$\\
\midrule
Full & \textbf{19.99} & \textbf{0.789} & \textbf{0.134} & \textbf{0.0039} & \textbf{0.0212} & \textbf{67.43}\\
w/o L1 & 19.51 & 0.772 & 0.141 & 0.0066 & 0.0354 & 67.05\\
w/o LPIPS & 19.86 & 0.780 & 0.145 & 0.0062 & 0.0337 & 66.03\\
w/o FFL & 19.85 & 0.786 & 0.135 & 0.0058 & 0.0308 & 66.64\\
\bottomrule
\end{tabular*}
\end{minipage}\par

\paragraph{Detector threshold.}
We vary the RMA threshold from 0.02 to 0.10 on UHDM. Our method has the lowest RMA among the compared methods at every tested threshold (Tab.~\ref{tab:threshold}). RMC averages continuous detector confidence and is independent of this threshold.
\begin{table}[!htb]
\centering
\caption{RMA at different detector thresholds on UHDM. Lower is better.}
\label{tab:threshold}
\papertablestyle
\begin{tabular*}{\linewidth}{@{\extracolsep{\fill}}lrrrrr@{}}
\toprule
Method & 0.02 & 0.03 & 0.05 & 0.07 & 0.10\\
\midrule
ESDNet-L & 0.1627 & 0.1193 & 0.0754 & 0.0534 & 0.0361\\
MZNet & 0.1425 & 0.1142 & 0.0808 & 0.0621 & 0.0454\\
MoCE-IR & 0.2203 & 0.1693 & 0.1110 & 0.0797 & 0.0531\\
OSEDiff & 0.0469 & 0.0355 & 0.0239 & 0.0174 & 0.0118\\
\midrule
\textbf{Ours} & \textbf{0.0440} & \textbf{0.0327} & \textbf{0.0212} & \textbf{0.0155} & \textbf{0.0106}\\
\bottomrule
\end{tabular*}
\end{table}

\section{Discussion}
Our method improves visible moir\'e suppression through a pretrained natural-image prior, but introduces trade-offs. Restored images may not align exactly with clean targets, lowering PSNR/SSIM, and severe color bands can retain color shifts. Two-stage inference is costly and not optimized for real-time use. We therefore assess fidelity, perceptual quality, and residual moir\'e jointly.

\section{Conclusion}
We present a content-preserving editing framework for real-world demoir\'eing. Endpoint supervision adapts a pretrained natural-image prior to restoration, while dual-scale condition-shift inference combines global correction with detail recovery. Moreover, we introduce DailyMoir\'e130 to evaluate challenging everyday screen photographs with color bands, fine stripes, and mixed-frequency artifacts. Our method achieves the lowest residual moir\'e area and strong human preference on this benchmark, with complementary gains on paired datasets. These results support pretrained editing priors for restoring moir\'e-corrupted content beyond the coverage of task-specific paired data.
\FloatBarrier
\label{paper:main-end}
\bibliographystyle{iclr2027_conference}
\bibliography{references}
\clearpage
\appendix
\raggedbottom
\section{Additional Qualitative Comparisons}\label{app:qualitative}
The following full-image comparisons show ten additional DailyMoir\'e130 examples without local magnification.
\par\addvspace{6pt}
\noindent\begin{minipage}{\linewidth}
\centering\captionsetup{hypcap=false}
\includegraphics[width=\linewidth]{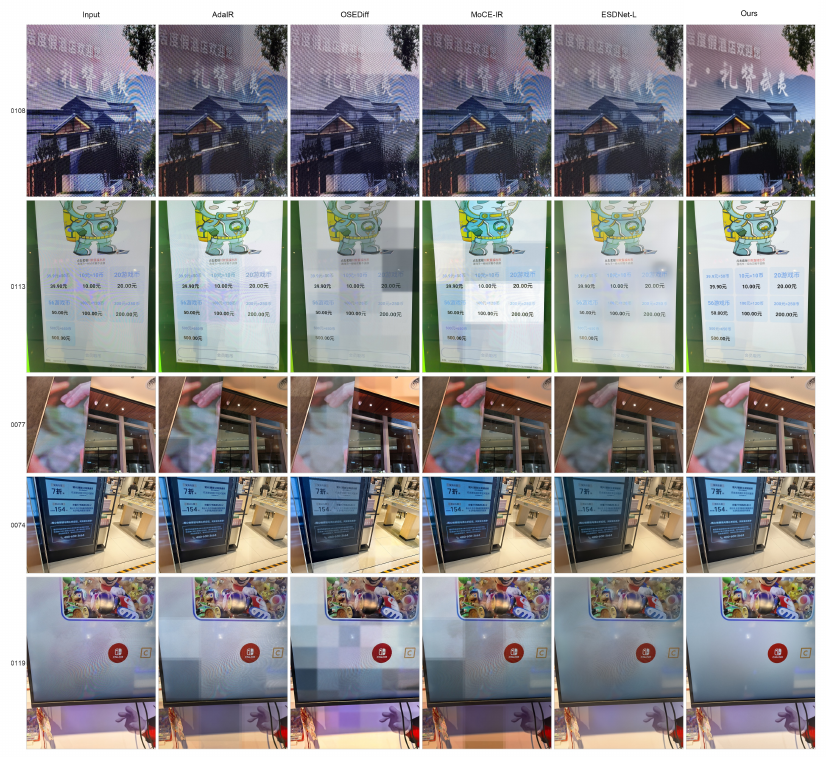}
\captionof{figure}{Additional full-image qualitative comparisons on DailyMoir\'e130, examples 1--5. Images are shown without local magnification.}\label{fig:supp-full1}
\end{minipage}\par\addvspace{6pt}
\par\addvspace{6pt}
\noindent\begin{minipage}{\linewidth}
\centering\captionsetup{hypcap=false}
\includegraphics[width=\linewidth]{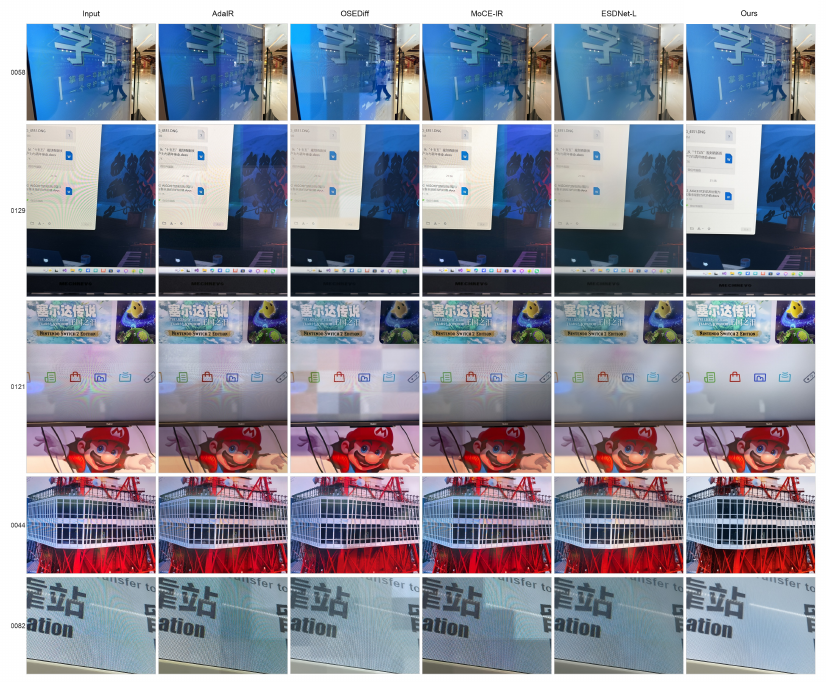}
\captionof{figure}{Additional full-image comparisons, examples 6--10. Each row compares Input, AdaIR, OSEDiff, MoCE-IR, ESDNet-L, and our method.}\label{fig:supp-full2}
\end{minipage}\par\addvspace{6pt}
\section{Endpoint-Supervised Training Objective}\label{app:loss}
The endpoint restoration loss in Eq.~\ref{eq:ltotal} is
\begin{equation}
\mathcal{L}_{\mathrm{restore}}=
\lambda_{\ell_1}\|\hat I-I_c\|_1+
\lambda_{\mathrm{lpips}}\,\mathrm{LPIPS}(\hat I,I_c)+
\lambda_{\mathrm{ffl}}\,\mathrm{FFL}(\hat I,I_c).
\label{eq:restore-detail}
\end{equation}
The weights are $\lambda_{\ell_1}=0.5$, $\lambda_{\mathrm{lpips}}=0.3$, and $\lambda_{\mathrm{ffl}}=0.2$, with $K=5$ unrolled steps. All three endpoint losses compare the prediction against the clean target. The total objective also retains the trajectory velocity loss in Eq.~\ref{eq:ltotal}.

The individual endpoint-loss ablations are reported in Sec.~\ref{sec:ablations} and Tab.~\ref{tab:loss}. Each variant retains velocity supervision and the other endpoint losses. Vanilla FM in Tab.~\ref{tab:ablation_component} instead uses standard single-timestep velocity supervision without endpoint losses, so that comparison changes both the training trajectory and the restoration objective.

\FloatBarrier
\par\noindent\begin{minipage}{\linewidth}
\section{Reproducibility Details}\label{app:recipe}
\subsection{Training recipe}
\begingroup
\captionsetup{hypcap=false}
\centering
\captionof{table}{Training recipe. Inference settings are specified below.}
\label{tab:training-recipe}
\papertablestyle
\begin{tabularx}{\linewidth}{@{}lY@{}}
\toprule
Item & Setting\\
\midrule
Base model & Qwen-Image-Edit-2509\\
Trainable parameters & LoRA adapters on the DiT backbone\\
Prompt & Remove moire patterns, color aliasing and color distortion from the image while preserving all details, colors, and textures of the original content.\\
Objective & Endpoint-supervised flow objective in Appendix~\ref{app:loss}\\
Data mixture & Public demoir\'eing data, document data, and self-reconstruction pairs\\
Training stream ratio & 4:1:1 at the manifest-stream level, comprising eight public/synthetic streams, two document streams, and two self-reconstruction streams\\
Public streams & FHDMi, UHDM, LDC-Moir\'e, MRBI, TIP-2018, WMD, UniDemoir\'e-style synthesis, and ScreenSim-style synthesis\\
Document streams & DocVQA English documents and Chinese document/form images\\
Self-reconstruction & Clean-image identity pairs from documents and UHDM\\
Batch / patch size & 96 / $256\times256$\\
Learning rate & Cosine schedule from $10^{-4}$ to $10^{-7}$\\
Fine-tuning length & 3k optimizer steps on four NVIDIA H100 GPUs\\
\bottomrule
\end{tabularx}
\par\endgroup\vspace{6pt}
The original prompt uses the spelling ``distoration'' rather than ``distortion''. Stream counts should not be interpreted as a measured per-image sampling ratio.
\end{minipage}\par

\subsection{Inference details}
We use the same prompt for all test images. Stage~1 resizes the short side to 1024 pixels, pairs the image with a black reference, and uses 10 denoising steps. Stage~2 performs one full-resolution loop from latent noise with $N=10$ steps and $r=0.3$. The first $\lceil rN\rceil=3$ steps use the upsampled Stage-1 edit latent, and the remaining seven use the original input latent.

\paragraph{Public implementation.}
Inference, training, and an interactive demo are available in the \href{https://github.com/Clausy9/DemoireEdit}{Demoir\'eEdit repository}; the restoration adapter is available on \href{https://huggingface.co/umu/DemoireEdit}{Hugging Face}. The maintained implementation additionally includes a latent-MSE extension, separate from the four-term objective reported in this manuscript. The released adapter is a historical checkpoint; its identity, provenance, and the limits of its correspondence to the maintained training recipe are documented in the repository model card.

\subsection{Runtime}
The full two-stage recipe takes approximately 260 seconds for a 4K image on a single NVIDIA H100 GPU in FP32.

\FloatBarrier
\par\noindent\begin{minipage}{\linewidth}
\section{DailyMoir\'e130 Dataset and Evaluation Protocol}\label{app:daily}
DailyMoir\'e130 contains 130 real-world screen photographs targeting challenging moir\'e in everyday life. The images cover signage, monitors, and laptops, with low-frequency color banding, fine high-frequency stripes, and mixed-frequency interference. Since these captures have no aligned clean targets, evaluation combines no-reference perceptual scores, residual moir\'e detector scores, and human preference (Tab.~\ref{tab:realdata}).

\paragraph{Public release.}
DailyMoir\'e130 is available on \href{https://huggingface.co/datasets/umu/DailyMoire130}{Hugging Face} (revision \href{https://huggingface.co/datasets/umu/DailyMoire130/tree/573da28262fecf5c14407dfa3fab9f080c3a15e1}{\texttt{573da28}}), with per-image checksums and an evaluation protocol. The public release retains the original resolutions and color profiles. It includes local blur in 18 images, while the remaining 112 retain their scene pixels; pixels outside the edited regions are unchanged. The localized edits are partial occlusion rather than comprehensive anonymization. Such edits can affect restoration outputs and no-reference metrics. The results reported in this manuscript use the historical inputs and have not been rerun on this public derivative.

For detector-based analysis, a restored image is divided into $P$ overlapping patches with moir\'e confidence $c_i\in[0,1]$. RMC averages these confidence scores, while RMA measures the fraction of patches exceeding 0.05, as defined in Eq.~\ref{eq:moire_metrics}.
\end{minipage}\par

\section{Human Preference Study}\label{app:studies}

\paragraph{Participants and inclusion.}
The blind study includes 31 participants who complete the randomized questionnaire and pass at least one of two embedded attention checks. Each questionnaire contains 18 Best-of-7 trials, two black-reference A/B trials, and two attention checks. Each attention check presents one intact restoration alongside six strongly corrupted variants. Among the 31 completed questionnaires, 29 participants pass both checks and two pass one. We also report the stricter 29-participant subset below. Incomplete questionnaires and responses from the earlier, non-randomized questionnaire version are excluded.

The study is conducted remotely on 13 laptops, 10 mobile devices, and eight desktops. A break is offered after Question 11. Summed active response time has a median of 10.5 minutes and an interquartile range of 6.9--13.8 minutes. Displays are not calibrated, and no formal visual-acuity or color-vision examination is imposed.

\paragraph{Image sampling and presentation.}
The schedule contains 40 counterbalanced templates of 18 distinct main-trial images, assigning each image five or six times. Public entry cycles through these templates. The completed cohort uses 24 distinct templates, with seven repeated assignments. Its 558 main trials cover all 130 DailyMoir\'e130 images, with a median of four evaluations per image and a range of one to nine.

Each trial compares the seven methods in Tab.~\ref{tab:realdata}, with identities hidden and randomized/counterbalanced across fixed A--G positions. Each method appears at each position 63--95 times. Participants inspect the input and candidates through full-image views and manually selected magnified crops. The same crop coordinates are used for every method and marked on the input overview.

\paragraph{Questionnaire criteria.}
For each image, participants answer three questions, translated here from the Chinese questionnaire.
\begin{itemize}[leftmargin=*,itemsep=2pt,topsep=3pt]
\item \textbf{Moir\'e suppression.} Which result has fewer residual moir\'e patterns, color bands, and screen-photography artifacts?
\item \textbf{Detail preservation.} Which result better preserves the text, edges, colors, and content details in the input, without obvious rewriting, missing content, or structural deformation?
\item \textbf{Overall quality.} Overall, which result has better and more natural visual quality?
\end{itemize}
Each question offers A--G, \emph{difficult to distinguish / no clear best}, and \emph{unable to answer / image display problem}. The detail question explicitly assesses preservation against the visible input.

\paragraph{Selection counts and uncertainty.}
Tab.~\ref{tab:study-counts} reports all 558 responses per criterion, including uncertain choices. These account for 3.0\%, 5.0\%, and 3.9\% of moir\'e, detail, and overall judgments. No main-trial response reports a display problem. A/B and attention trials are excluded.

\begin{table}[!htb]
\centering
\caption{Best-of-7 selection counts on DailyMoir\'e130. Each column contains 558 responses, including uncertain responses. Percentages are reported in Tab.~\ref{tab:realdata}.}
\label{tab:study-counts}
\papertablestyle
\begin{tabular*}{\linewidth}{@{\extracolsep{\fill}}lrrr@{}}
\toprule
Method / response & Moir\'e suppression & Detail preservation & Overall quality\\
\midrule
Ours & 417 & 382 & 398\\
ESDNet-L & 48 & 48 & 49\\
Uni-control & 30 & 38 & 37\\
MZNet & 15 & 16 & 14\\
OSEDiff & 13 & 19 & 15\\
MoCE-IR & 10 & 13 & 12\\
AdaIR & 8 & 14 & 11\\
Uncertain & 17 & 28 & 22\\
\midrule
Total & 558 & 558 & 558\\
\bottomrule
\end{tabular*}
\end{table}

We use 20,000 bootstrap resamples and percentile 95\% confidence intervals. The primary analysis samples 31 participants with replacement, retaining all their trials. An image-level sensitivity analysis separately samples 130 images with replacement, retaining all their responses. Each resample divides pooled selections by pooled trials, including uncertain responses. These are separate one-way cluster analyses. Tab.~\ref{tab:study-ci} reports both intervals for our method and ESDNet-L, the most-preferred baseline on all three criteria.

\begin{table}[!htb]
\centering
\caption{Best-of-7 selection rates and separate cluster-bootstrap 95\% confidence intervals (\%). Participant clustering is the primary analysis, and image clustering is a sensitivity analysis.}
\label{tab:study-ci}
\papertablestyle
\setlength{\tabcolsep}{3pt}
\begin{tabular*}{\linewidth}{@{\extracolsep{\fill}}llrrr@{}}
\toprule
Criterion & Method & Rate & Participant CI & Image CI\\
\midrule
Moir\'e suppression & Ours & 74.7 & [70.4, 78.9] & [69.8, 79.3]\\
& ESDNet-L & 8.6 & [6.1, 11.3] & [6.0, 11.4]\\
Detail preservation & Ours & 68.5 & [62.9, 73.7] & [63.4, 73.4]\\
& ESDNet-L & 8.6 & [5.9, 11.5] & [5.9, 11.6]\\
Overall quality & Ours & 71.3 & [65.1, 76.9] & [66.3, 76.2]\\
& ESDNet-L & 8.8 & [5.7, 12.2] & [6.1, 11.8]\\
\bottomrule
\end{tabular*}
\end{table}

\paragraph{Attention-check sensitivity.}
Requiring both checks retains 29 participants and 522 trials per criterion. Our method receives 74.3\% moir\'e, 69.9\% detail, and 73.2\% overall preference, compared with 9.0\%, 8.4\%, and 8.0\% for ESDNet-L. The preference advantage persists under this stricter rule.

Our method receives the unique highest overall-quality vote count for 30 of 31 participants. The three criteria select the same outcome in 479/558 trials (85.8\%), so they are not independent replications. The study supports perceived detail preservation under remote viewing, while unpaired images do not establish exact recovery of details hidden by corruption.

\FloatBarrier

\par\noindent\begin{minipage}{\linewidth}
\section{Native Qwen-Edit Negative Control}\label{app:native}
We also test native Qwen-Image-Edit-2509 without demoir\'e-specific LoRA training. Zero-shot attempts with the same prompt can turn structured moir\'e into large color blocks instead of preserving source content. This control separates the pretrained editing prior from task adaptation; it does not by itself establish the individual necessity of every loss or inference component.
\par\vspace{6pt}
\centering\captionsetup{hypcap=false}
\includegraphics[width=.9\linewidth]{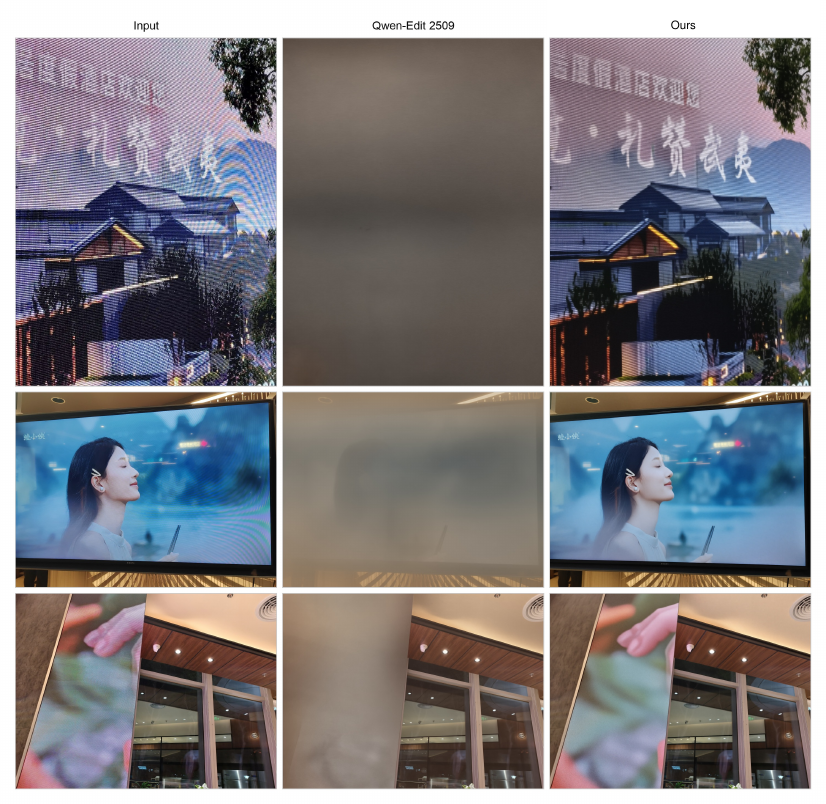}
\captionof{figure}{Native Qwen-Image-Edit-2509 negative control. The illustrated zero-shot outputs have large color-block artifacts and altered structure.}\label{fig:native}
\end{minipage}\par

\FloatBarrier
\section{Additional paired-benchmark metrics}\label{app:paired-metrics}
Tab.~\ref{tab:paired-additional} completes the paired comparison in Tab.~\ref{tab:main_paired}. Our method achieves the highest MANIQA on both datasets. MZNet achieves the lowest LPIPS on both datasets and the highest SSIM on FHDMi, while MoCE-IR achieves the highest SSIM on UHDM. Together with PSNR in the main table, these results characterize the trade-off between full-reference fidelity and residual moir\'e suppression.

\begin{table}[!ht]
\centering
\caption{Additional metrics on FHDMi and UHDM. Best in \textbf{bold}, second-best \underline{underlined}. Uni-control denotes UniDemoir\'e-style augmentation of ESDNet-L.}
\label{tab:paired-additional}
\papertablestyle
\setlength{\tabcolsep}{2pt}
\begin{tabular*}{\linewidth}{@{\extracolsep{\fill}}lrrrrrr@{}}
\toprule
& \multicolumn{3}{c}{FHDMi} & \multicolumn{3}{c}{UHDM}\\
\cmidrule(lr){2-4}\cmidrule(l){5-7}
Method & SSIM$\uparrow$ & LPIPS$\downarrow$ & MANIQA$\uparrow$ & SSIM$\uparrow$ & LPIPS$\downarrow$ & MANIQA$\uparrow$\\
\midrule
Input & .707 & .265 & .311 & .560 & .335 & .380\\
ESDNet-L & .789 & .183 & .315 & \underline{.837} & \underline{.118} & .411\\
MoCE-IR & \underline{.831} & \underline{.151} & .307 & \textbf{.838} & .145 & .370\\
AdaIR & .808 & .180 & .299 & .815 & .173 & .358\\
OSEDiff & .739 & .201 & \underline{.340} & .784 & .205 & .400\\
MZNet & \textbf{.836} & \textbf{.109} & .325 & .835 & \textbf{.108} & .413\\
Uni-control & .793 & .182 & .313 & .831 & .126 & \underline{.414}\\
\midrule
\textbf{Ours} & .785 & \underline{.151} & \textbf{.363} & .789 & .134 & \textbf{.419}\\
\bottomrule
\end{tabular*}
\end{table}

\FloatBarrier
\paragraph{Content and color fidelity.}
We further evaluate content fidelity against the clean references on UHDM using TOPIQ-FR~\citep{chen2024topiq}, PieAPP~\citep{pieapp2018}, and AHIQ~\citep{ahiq2022}, together with CIE chroma and hue differences~\citep{ciede2000}. TOPIQ-FR uses the clean reference, whereas TOPIQ in Tab.~\ref{tab:main_paired} is the no-reference variant. Tab.~\ref{tab:content} shows that our method achieves the highest TOPIQ-FR and the second-best PieAPP and AHIQ among the compared methods. Its chroma and hue errors are comparable to those of the feed-forward restorers, with MZNet achieving the lowest errors. These paired measurements complement the dedicated human detail assessment on DailyMoir\'e130, although neither establishes exact recovery of details that are unobservable in the corrupted input.

\begin{table}[!ht]
\centering
\caption{Content and color fidelity on UHDM. TOPIQ-FR is full-reference, whereas TOPIQ in Tab.~\ref{tab:main_paired} is no-reference. Best values are bold. Uni-control denotes UniDemoir\'e-style augmentation of ESDNet-L.}
\label{tab:content}
\papertablestyle
\begin{tabular*}{\linewidth}{@{\extracolsep{\fill}}lrrrrr@{}}
\toprule
Method & TOPIQ-FR$\uparrow$ & PieAPP$\downarrow$ & AHIQ$\uparrow$ & $|\Delta C^*|\downarrow$ & $\Delta H^*\downarrow$\\
\midrule
ESDNet-L & 0.639 & 1.137 & 0.433 & 4.454 & 2.776\\
Uni-control & 0.641 & 1.045 & \textbf{0.439} & 4.273 & 2.865\\
MZNet & 0.664 & 0.991 & 0.431 & \textbf{3.995} & \textbf{2.407}\\
MoCE-IR & 0.658 & \textbf{0.964} & 0.433 & 4.322 & 2.727\\
OSEDiff & 0.588 & 1.310 & 0.284 & 6.352 & 4.034\\
\midrule
\textbf{Ours} & \textbf{0.670} & 0.984 & 0.435 & 4.422 & 2.643\\
\bottomrule
\end{tabular*}
\end{table}

\end{document}

%% file: authors.tex
\author{%
Shiyang Zhou\textsuperscript{1,2}\quad
Haijin Zeng\textsuperscript{1,*}\quad
Congpei Qiu\textsuperscript{2,3}\quad
Zhengcen Li\textsuperscript{1}\quad
Ping Chen\textsuperscript{1}\\
\bfseries Xiangming Wang\textsuperscript{1}\quad
Zheng Li\textsuperscript{1}\quad
Yongyong Chen\textsuperscript{1}\quad
Jingyong Su\textsuperscript{1,*}\\[6pt]
{\normalfont\textsuperscript{1}Harbin Institute of Technology, Shenzhen}\\
{\normalfont\textsuperscript{2}Shenzhen Loop Area Institute\quad
\textsuperscript{3}Xi'an Jiaotong University}\\[3pt]
{\normalfont\textsuperscript{*}Corresponding authors.}%
}